\documentclass[trsc,nonblindrev]{informs4} 

\OneAndAHalfSpacedXI 

\usepackage{natbib}
 \bibpunct[, ]{(}{)}{,}{a}{}{,}%
 \def\bibfont{\small}%

\usepackage{hyperref}

\usepackage[nolist]{acronym}
\usepackage{mathrsfs}
\usepackage{dirtytalk}
\usepackage{multirow}
\usepackage{booktabs}
\usepackage{graphicx}
\usepackage{subcaption}
\usepackage{array}
\usepackage{tabularx}
\usepackage{float}

\usepackage{rotating}

\begin{acronym}
    \acro{NS}{nucleus sampling}
    \acro{GS}{greedy sampling}
    \acro{TSP}{Traveling Salesman Problem}
    \acro{VRP}{Vehicle Routing Problem}
    \acro{CVRP}{Capacitated Vehicle Routing Problem}
    \acro{OR}{Operations Research}
    \acro{BKS}{best-known solution}
    \acro{AM}{Attention Model}
    \acro{CO}{Combinatorial Optimization}
    \acro{MILP}{Mixed Integer Linear Programming}
    \acro{CV}{Computer Vision}
    \acro{NLP}{Natural Language Processing}
    \acro{CNN}{Convolutional Neural Network}
    \acro{ML}{Machine Learning}
    \acro{HGS}{Hybrid Genetic Search}
    \acro{LNS}{Large Neighborhood Search}
    \acro{GA}{Genetic Algorithm}
    \acro{CW}{Clarke-Wright}
    \acro{LK}{Lin-Kernighan}
    \acro{LKH}{Lin-Kernighan-Helsgaun}
    \acro{LKH-3}{Lin-Kernighan-Helsgaun-3}
    \acro{LLaMA-2}{Large Language Model Meta AI 2}
    \acro{DL}{Deep Learning}
    \acro{PN}{Pointer Network}
    \acro{Seq2Seq}{Sequence-to-Sequence}
    \acro{PPO}{Proximal Policy Optimization}
    \acro{LLM}{Large Language Model}
    \acro{BFS}{Breadth-First Search}
    \acro{MCTS}{Monte Carlo Tree Search}
    \acro{AI}{Artificial Intelligence}
    \acro{GPT}{Generative Pre-trained Transformer}
    \acro{ChatGPT}{Chat Generative Pre-trained Transformer}
    \acro{BCP}{Branch-Cut-and-Price}
    \acro{lm-SRC}{Limited Memory Subset Row Cut}
    \acro{SA}{Simulated Annealing}
    \acro{PILS}{Pattern Injection Local Search}
    \acro{KGLS}{Knowledge Guided Local Search}
    \acro{RL}{Reinforcement Learning}
    \acro{POPMUSIC}{Partial Optimization Metaheuristic under Special Intensification Conditions}
    \acro{SISR}{Slack Induction by String Removal}
    \acro{SIFT}{Scale Invariant Feature Transform}
    \acro{HOG}{Histogram of Oriented Gradients}
    \acro{GNN}{Graph Neural Network}
    \acro{NN}{Neural Network}
    \acro{GELU}{Gaussian Error Linear Unit}
    \acro{RELU}{Rectified Linear Unit}
    \acro{T5}{Text-to-Text Transfer Transformer}
    \acro{LM}{Language Model}
    \acro{Prefix LM}{Prefix Language Model}
    \acro{SGD}{Stochastic Gradient Descent}
    \acro{CL}{Curriculum Learning}
    \acro{MITSC}{MIT SuperCloud}
    \acro{MCVRP}{Montreal Capacitated Vehicle Routing Problem}
    \acro{MLOps}{Machine Learning Operations}
    \acro{RNN}{Recurrent Neural Network}
    \acro{MHA}{Multi-Head Attention}
    \acro{VRP-T5}{VRP Text-to-Text Transfer Transformer}
    \acro{FTM-VRP}{Foundational Transformer Model for the Vehicle Routing Problem}
    \acro{FM-MCVRP}{Foundation Model for the Montreal Capacitated Vehicle Routing Problem}
    \acro{MP}{Mathematical Programming}
    \acro{DRL}{Deep Reinforcement Learning}
\end{acronym}

\TheoremsNumberedThrough     
\ECRepeatTheorems

\EquationsNumberedThrough    

\MANUSCRIPTNO{TBD}

\begin{document}



\RUNTITLE{Foundation Model for the Montreal Capacitated Vehicle Routing Problem}

\TITLE{Learning to Deliver: a Foundation Model for the Montreal Capacitated Vehicle Routing Problem}

\ARTICLEAUTHORS{%
\AUTHOR{Samuel J. K. Chin, Matthias Winkenbach}
\AFF{
Massachusetts Institute of Technology, Center for Transportation \& Logistics, 77 Massachusetts Ave, Cambridge, MA 02139,
\EMAIL{jkschin@mit.edu},
\EMAIL{mwinkenb@mit.edu}
}
\AUTHOR{Akash Srivastava}
\AFF{
MIT-IBM Watson AI Lab, 314 Main St, Cambridge, MA 02142,
\EMAIL{Akash.Srivastava@ibm.com}
}
} 

\ABSTRACT{%
In this paper, we present the \ac{FM-MCVRP}, a novel \ac{DL} model that approximates high-quality solutions to a variant of the \ac{CVRP} that characterizes many real-world applications.
The so-called \ac{MCVRP}, first formally described by \cite{Bengio2021MachineDhorizon}, is defined on a fixed and finite graph, which is analogous to a city.
Each \ac{MCVRP} instance is essentially the sub-graph connecting a randomly sampled subset of the nodes in the fixed graph, which represent a set of potential addresses in a real-world delivery problem on a given day.
Our work exploits this problem structure to frame the \ac{MCVRP} as an analogous \ac{NLP} task.
Specifically, we leverage a Transformer architecture embedded in a \ac{LLM} framework to train our model in a supervised manner on computationally inexpensive, sub-optimal \ac{MCVRP} solutions obtained algorithmically.
Through comprehensive computational experiments, we show that \ac{FM-MCVRP} produces better \ac{MCVRP} solutions than the training data and generalizes to larger sized problem instances  not seen during training.
Even when compared to near-optimal solutions from state-of-the-art heuristics, \ac{FM-MCVRP} yields competitive results despite being trained on inferior data.
For instance, for 400-customer problems, \ac{FM-MCVRP} solutions on average fall within 2\% of the benchmark.
Our results further demonstrate that unlike prior works in the literature, \ac{FM-MCVRP} is a unified model, which performs consistently and reliably on a range of problem instance sizes and parameter values such as the vehicle capacity.
}%



\maketitle

%



\acresetall
\section{Introduction}
\label{sec:introduction}

The \ac{CVRP} is one of the most well-researched \ac{CO} problems in the \ac{OR}, transportation, and logistics literature, given its many commercial and non-commercial applications.
To this day, it continues to be an active area of academic research with advances being made in two main directions.
First, researchers seek to develop new methodological approaches either to solve existing variants of the problem more efficiently, and for larger problem instances, or to uncover new best known solutions to benchmark problem instances from the literature \citep{Uchoa2017NewProblem, Arnold2019EfficientlyProblems}.
Second, inspired by increasingly sophisticated routing challenges faced by the transportation and logistics industry, authors define and investigate new and more advanced variants of the problem to capture additional constraints and nuanced problem characteristics \citep[cf.,][]{Vidal2020AVariants}.

Methodologically, the \ac{OR} literature on the \ac{CVRP} and its many variants can be grouped into 
(i) exact methods, including \ac{MP},
(ii) heuristic approaches, and
(iii) metaheuristic approaches.
The dominant exact method to solve \ac{CO} problems such as the \ac{CVRP} is \ac{MP}. 
However, the NP-hard nature of these problems \citep[cf.,][]{Cook2011TheProblem} quickly renders such models intractable for realistically sized problem instances. 
Therefore, for solving large-scale problem instances often faced in real industrial applications, most state-of-the-art solution methods rely on heuristic and metaheuristic approaches \citep{Arnold2019Knowledge-guidedProblem, Vidal2022HybridNeighborhood}.
Most notably, the \ac{LKH-3} heuristic by \cite{Helsgaun2017AnProblems} and \ac{HGS} \citep{Vidal2022HybridNeighborhood} are currently among the best performing solution approaches to the \ac{CVRP} \citep[cf., Table 5 in ][]{Vidal2022HybridNeighborhood}.

\ac{ML} methods, on the other hand, dominated the fields of \ac{CV} and \ac{NLP} in recent years. 
Further, \ac{ML} has found numerous highly visible applications -- from strategic game playing \citep[see, e.g.,][]{Silver2018ASelf-play} to molecular biology \citep[see, e.g.,][]{Jumper2021HighlyAlphaFold}.
More recently, publicly discussed generative \ac{AI} models such as \ac{ChatGPT} have enabled human-like conversations with a machine.
The rapidly advancing algorithmic performances are constantly pushing the boundaries of achievable state-of-the-art results.
As these problems in general have an extremely large state space, a natural next question in research is to what extent \ac{ML} can be applied to \ac{CO} problems and complement or even replace existing \ac{OR} methods \citep{Bengio2021MachineDhorizon}.
Specifically, \acp{PN}, first proposed by \cite{Vinyals2015PointerNetworks}, demonstrate how a \ac{NN} can approximate solutions to the \ac{TSP}. 
This work has since inspired numerous approaches to solving routing problems with \ac{DL} \citep[see, e.g.,][]{Nazari2018ReinforcementProblem, Lu2019AProblems, Kool2019AttentionProblems, Kwon2020Pomo:Learning}.
The idea of leveraging \acp{NN} to solve \ac{CO} problems such as the \ac{TSP} is not entirely new though.
For instance, \citet{Hopfield1985NeuralProblems} were the first to solve small \ac{TSP} instances with a so-called Hopfield network.

\subsection{Contributions of This Work}

There are four main contributions of this work.
First, we show that the recent methodological advances in the field of \acp{LLM} can be applied successfully to solving combinatorial optimization problems in the transportation and logistics domain, such as vehicle routing.
Specifically, we propose the \ac{FM-MCVRP}, a novel supervised \ac{DL} model that approximates near-optimal solutions to the \ac{MCVRP}, a particularly relevant variant of the \ac{CVRP}, which mimics many real-world delivery problems \citep[cf.,][]{Bengio2021MachineDhorizon}.
Unlike recent work from \citet{Kool2019AttentionProblems} and \citet{Kwon2020Pomo:Learning} who build on the Transformer architecture \citep{Vaswani2017AttentionNeed} in a \ac{PN} framework, \ac{FM-MCVRP} is to the best of our knowledge the first \ac{DL} model leveraging the Transformer architecture in an \ac{LLM} framework for routing problems.
Building on the \ac{T5} model by \cite{Raffel2020ExploringTransformer}, which has been largely applied to \ac{NLP} problems, we model the underlying routing problem as an analogous \ac{NLP} task that predicts the nodes to visit defined on a fixed graph instead of using a \ac{PN}.

Second, our proposed model is a unified model that accepts a range of problem sizes and vehicle capacities. This makes our model highly applicable to real-world problems instead of only performing well on stylized problem instances with tightly controlled parameters. 
Previous works, such as \citet{Kool2019AttentionProblems} and \citet{Kwon2020Pomo:Learning}, train a single model for each given combination of problem size and vehicle capacity, which seems impractical for real-world operations as companies do not have the capability to manage a large number of \ac{ML} models that are trained for each possible combination of problem size and capacity they might encounter.
Moreover, we can show that our proposed model can solve up to 800-node problem instances, compared to \citet{Kool2019AttentionProblems} and \citet{Kwon2020Pomo:Learning} who solve up to 100-node problem instances as larger instances result in divergence in the training process (see Appendix \ref{app:retraining-am}).
However, many real-world delivery problems, especially in last-mile logistics, contain more than 100 customers.

Third, we show that our proposed model is able to outperform the algorithmically obtained solutions it was trained on.
This finding has strong managerial implications.
Given the size and complexity of many real-world routing problems, and due to a lack of expertise and ability to invest into high-performing algorithms explicitly tailored towards their specific problem, many companies rely on route planning tools based on relatively simple algorithms, decision rules, and human experience.
Hence, many companies continue to amass large quantities of potentially sub-optimal route sequences that are obtained either algorithmically or by observing how routes are actually executed by drivers.
Since the \ac{FM-MCVRP} is agnostic to the method used to obtain the solutions on which it is trained, it allows companies to improve route quality by leveraging existing routing data. 
This means that they can gradually improve the quality of their routes by learning from readily available data without having to invest into designing and maintaining an explicit algorithm tailored to their business needs.
Notably, our model is able to find solutions that are competitive to and may even outperform state-of-the-art benchmark methods on certain problem instances, despite being trained on inferior, yet computationally inexpensive solutions.

Fourth, we show that \ac{FM-MCVRP} generalizes well to larger problem instances.
Specifically, we train our model on 20 to 400-node problem instances before testing it on 600 and 800-node problem instances.
We show that \ac{FM-MCVRP} is able to consistently maintain its favorable performance characteristics for these larger problem instances. 
The ability to generalize to larger, and previously unseen instance sizes while maintaining a high solution quality has important implications for many real-world routing problems.
The number of stops on a route may vary widely depending on the commercial, geographical, and operational context (e.g., product and customer characteristics, stop density, vehicle technology, etc.). 
While a company's training data may be skewed towards certain instance sizes, the trained \ac{FM-MCVRP} can still yield good results for previously unseen or less prominent instance sizes.
Similarly, a company may be able to train \ac{FM-MCVRP} on historical data from routes it currently performs and leverage the trained model to plan routes even as the demand profile of it clients or the operational and technological context of its routes changes over time.

The remainder of this paper is structured as follows.
In Section \ref{sec:lit}, we review relevant literature for solving or approximating solutions to the \ac{CVRP}.
We cover exact methods, heuristics and metaheuristics, \ac{DL} approaches, and finally decoding strategies for \ac{DL} approaches.
In Section \ref{sec:problem}, we formally define the \ac{CVRP} and revisit the formal definition of the \ac{MCVRP} presented by \cite{Bengio2021MachineDhorizon}.
In Section \ref{sec:method}, we detail our methodology and present our \ac{FM-MCVRP} model.
In Section \ref{sec:experiment_setup}, we describe the setup of our computational experiments.
The results and insights from these experiments are presented in Section \ref{sec:results}.
In Section \ref{sec:managerial_insights}, we provide the managerial insights.
We conclude with a summary of our work and an overview of what we consider fruitful directions for future research in Section \ref{sec:conclusion}.

\section{Related Literature}
\label{sec:lit}

In the following, we provide an overview of current state-of-the-art methods to solve the \ac{CVRP} and related problems both from an \ac{OR} and from an \ac{ML} perspective.
To frame the discussion of \ac{CVRP} heuristics, we follow the definition from \cite{Toth2002TheProblem} which broadly distinguishes three categories: construction, two-phase, and improvement methods.
Two-phase methods can be further classified into \emph{cluster-first, route-second} methods and \emph{route-first, cluster-second} methods.
The performance of the algorithms within each category differ and it is important to be cognizant of the class to which an algorithm belongs to and compare algorithms within its class.
Given the nature of our proposed method, we focus our literature review on construction methods.

We first cover the best performing exact methods in Section \ref{sec:lit_exact} before providing an overview of state-of-the-art heuristic and metaheuristic approaches in Section \ref{sec:lit_heur}.
In Section \ref{sec:lit_ML}, we then survey the most promising \ac{DL} approaches to common routing problems. 
Section \ref{sec:lit_dec} provides additional insights into state-of-the-art decoding strategies, which are a critical determinant of the performance of learning-based approaches to \ac{CO} problems. In Section \ref{sec:lit_gap}, we distill the research gap that this paper is intended to fill.

\subsection{Exact Methods}
\label{sec:lit_exact}
Exact methods in \ac{OR} primarily involve \ac{MP}, and so-called matheuristics, which integrate \ac{MP} into metaheuristic frameworks for enhanced problem-solving efficiency.
In this section, we first discuss \ac{MP} approaches, followed by matheuristic approaches.
A recent advancement in \ac{MP} approaches to finding exact solutions to the \ac{CVRP} has been the introduction of \ac{BCP} by \cite{Pecin2017ImprovedRouting}.
The authors combine a myriad of innovations previously proposed in the literature on branch-and-price methods with so-called \acp{lm-SRC}, which enable significant efficiency gains in the pricing subproblem.
They use their method to efficiently solve \ac{CVRP} instances of up to 360 customers, which is a relevant instance size for many real-world industry applications.
\cite{Pessoa2020AProblems} further extend the work of \cite{Pecin2017ImprovedRouting} for a more general definition of the \ac{VRP}.
For a recent in-depth review of exact \ac{BCP} methods, we refer the reader to \cite{Costa2019ExactRouting}.

Matheuristics on the other hand combine \ac{MP} and metaheuristic approaches (see Section \ref{sec:lit_heur}). 
\cite{Queiroga2021AProblem} propose a matheuristic coined \ac{POPMUSIC} for the \ac{CVRP}.
Using best known solutions from the \ac{CVRP} benchmark instances of \cite{Uchoa2017NewProblem} as an initial solution, their method is able to find new best known solutions for several instances by breaking them down into subproblems that are solved using a modified \ac{BCP} algorithm as a heuristic.
More recently, \cite{Skalnes2023AProblem} propose another state-of-the-art matheuristic.
The authors use a construction heuristic to create an initial solution which, in turn, is refined by an improvement heuristic. 
The refined solution then serves as a starting solution for an exact branch-and-cut algorithm.
The improvement and branch-and-cut steps are repeated until an optimal solution is found or another termination criteria is met.
The method of \cite{Skalnes2023AProblem} is able to obtain the currently best known solutions for \ac{CVRP} instances with more than 10,000 customers from the dataset of \cite{Arnold2019EfficientlyProblems}.
For a comprehensive review of prior works on matheuristics, we refer the reader to \cite{Archetti2014AProblems}.

\subsection{State-of-the-Art Heuristics and Metaheuristics}
\label{sec:lit_heur}

The \ac{CVRP} is an NP-hard problem rendering large-scale problem instances that correspond to many real-world applications intractable.
Therefore, efficient approximate solution techniques are required.
State-of-the-art approaches to solving large \ac{CVRP} instances frequently involve a local search heuristic incorporated in a metaheuristic framework \citep{Vidal2022HybridNeighborhood}.
As a \ac{CVRP} solution consists of multiple \ac{TSP} tours, \ac{TSP} local search heuristics are commonly employed in solving \acp{CVRP}.
Commonly employed \ac{TSP} local search heuristics include the 2-OPT, 3-OPT \citep{Croes1958AProblems}, the \ac{LK} heuristic \citep{Lin1973AnProblem} and the \ac{LKH} heuristic \citep{Helsgaun2000AnHeuristic}. 
Note that \ac{LK} and \ac{LKH} are the generalizations of the 2-OPT and 3-OPT local search operator to the $k$-opt operator.
As \ac{LKH} is only applicable to the \ac{TSP}, \cite{Helsgaun2017AnProblems} propose \ac{LKH-3}, an extension of \ac{LKH} specifically for \acp{CVRP}.
Common metaheuristic frameworks that have been successfully applied to the \ac{CVRP} are \ac{SA} \citep{Bertsimas1993SimulatedAnnealing}, \acp{GA} \citep{Holland1992AdaptationIntelligence}, and \acf{LNS} \citep{Shaw1998UsingProblems}.

\ac{HGS} \citep{Vidal2022HybridNeighborhood} is a current state-of-the-art metaheuristic that is predominantly an improvement method. 
It derives its remarkable performance from integrating a local search heuristic (SWAP*), which is essentially a classic \emph{swap} operator without an insertion in place, in a \ac{GA} framework.
The \ac{HGS} algorithm maintains a pool of feasible and infeasible solutions at all times. 
Following a typical \ac{GA} logic, it first selects two solutions (genes) from this pool and combines the two solutions with an ordered crossover \citep{Oliver1987AProblem} to obtain a new solution. 
The algorithm then conducts a controlled neighborhood search that explores both feasible and infeasible solutions to find a new local minimum.
In the event where the resulting solution continues to be infeasible, a repair operator is applied with 50\% probability. After that, the solution is either added to the feasible or the infeasible solution pool.

\ac{KGLS} \citep{Arnold2019Knowledge-guidedProblem} and \ac{SISR} \citep{Christiaens2020SlackProblems} are two other metaheuristic approaches that achieve competitive results \citep[cf., Table 5 and 6 in ][]{Vidal2022HybridNeighborhood} on the benchmark instances from \cite{Uchoa2017NewProblem}.
\ac{KGLS} relies on penalization of arcs during the local search process.
The penalties are computed based on features that were obtained from studying the common characteristics of good and bad solutions.
\ac{SISR} proposes a \emph{destroy}, a \emph{repair}, and a fleet minimization procedure, where the \emph{destroy} operator has a novel property of spatial slack and the \emph{repair} operator is categorized as a greedy insertion with blinks \citep{Christiaens2020SlackProblems}.

Among the heuristics and metaheuristics discussed in this section, \ac{HGS} is the highest performing in terms of solution quality obtained on the benchmark instances of \cite{Uchoa2017NewProblem} under a fixed time limit \citep[cf., Table 5 and 6 in ][]{Vidal2022HybridNeighborhood}. 

\subsection{Deep Learning Approaches}
\label{sec:lit_ML}
Most recently discussed advances in \ac{AI}, such as \ac{ChatGPT} and others, are built on \ac{DL} methods.
In this section, we briefly provide a few definitions to help the reader make the translation from terms and concepts originally developed in the context of other \ac{DL} applications to the context of the \ac{CVRP}.
We then introduce the \emph{Attention} mechanism and the seminal \emph{Transformer} \citep{Vaswani2017AttentionNeed} that forms the basis of almost all state-of-the-art \ac{NLP} and \ac{CV} methods currently discussed.
Following that, we introduce two other main \ac{NLP} architectures: the \ac{LM} architecture and the \ac{Prefix LM} architecture.
Finally, we discuss recent state-of-the-art \ac{DL} approaches to solving the \ac{CVRP}, and their relative performance compared to state-of-the-art heuristics and metaheuristics.

\paragraph{Definitions.}
In the \ac{DL} literature, the term \emph{embedding} refers to a vector representation of an object transformed from its original features.
In the context of \ac{NLP}, \ac{CV}, and the \ac{CVRP}, the object corresponds to a word, image, or node, respectively.
In routing, a node could be expressed as a vector representing its $(x,y)$-coordinates, demand, and other relevant features.
Note that while two closely co-located nodes with different demand quantities can be considered dissimilar when simply using a norm over its original features, transforming these nodes into their embeddings may uncover more deeply rooted similarities in their node characteristics, making them appear more similar to one another in the specific context of the underlying routing problem.

In the \ac{NLP} literature \citep[see, e.g., ][]{Vaswani2017AttentionNeed}, a \emph{token} is generally analogous to a word in a sentence. 
In the context of routing, a token corresponds to a node to be visited at a specific position in a given route sequence.

\paragraph{Attention and Transformers.}
So-called \emph{Attention} is now a commonly used mechanism in \ac{DL} that was first introduced by \cite{Bahdanau2014NeuralTranslate}, which selectively places emphasis on specific parts of the input sequence.
In their seminal \emph{Transformer} paper, \cite{Vaswani2017AttentionNeed} propose an extension to this mechanism (see Appendix \ref{app:attention}), which we are referring to in this paper. 
We can further distinguish two variants of the attention mechanism:
\emph{fully visible} attention, where all tokens are able to \emph{attend} to each other, and
\emph{causal} attention \citep[see][]{Raffel2020ExploringTransformer}, where tokens in earlier parts of a sequence are unable to \emph{attend} to tokens in later parts of the sequence.
Finally, the Attention mechanism can be augmented with \ac{MHA}, which applies the Attention mechanism with different learned weight matrices \citep{Vaswani2017AttentionNeed}.

The key building blocks of a Transformer model are an \emph{encoder} and a \emph{decoder} and thus it is also commonly known as the encoder-decoder architecture.
The \emph{encoder} is  made up of a series of encoder blocks, which themselves contain \ac{MHA} and feed-forward layers.
Essentially, it is a function that takes as input a matrix of features and transforms the matrix into embeddings.
The \emph{decoder}, in turn, consists of a series of decoder blocks, which themselves contain \ac{MHA}, causal \ac{MHA} and feed-forward layers.
It is a function that takes the encoder embeddings and node features in the partial solution as input and outputs nodes in an autoregressive manner.
Expressed in the context of vehicle routing, the encoder learns a representation of all nodes in the underlying graph of the \ac{MCVRP} and the decoder seeks to capture the distribution of solutions in the graph.

\paragraph{NLP model architectures.}
The field of \ac{NLP} broadly consists of three main model architectures: the encoder-decoder as described above, the \ac{LM}, and the \ac{Prefix LM}.
The \ac{LM} architecture, which is commonly known as the decoder-only architecture, is first proposed in the \ac{GPT} model by \cite{Radford2019LanguageLearners}.
In this model, \emph{causal} attention is applied to the model such that a token at any given position can only see the previous tokens in the input and not future tokens.

The \ac{Prefix LM} is essentially a modified version of the decoder-only  architecture. Here, the model has a prefix section where all tokens have \emph{fully visible} attention.
For more details, we refer the reader to \cite{Raffel2020ExploringTransformer}.
In the context of a \ac{MCVRP}, the classic encoder-decoder and \ac{Prefix LM} are the most relevant architectures, as the problem is defined on a fully connected graph in which all nodes are fully visible (i.e., connected) to each other.
While this is consistent with an encoder-decoder or a \ac{Prefix LM} architecture, the causal mask in an \ac{LM} architecture would limit the nodes' visibility of each other.

\paragraph{Deep Learning for routing problems.}
\cite{Vinyals2015PointerNetworks} are the first authors to present \ac{DL} methods that attempt to approximate solutions to the \ac{TSP}.
Their method, \acp{PN}, is a \ac{RNN} based method inspired by \ac{Seq2Seq} models in \ac{NLP}.
It solves the \ac{TSP} by taking an input sequence of nodes and outputs a sequence of nodes through attending to the input nodes.

\citet{Kool2019AttentionProblems} connect the the idea of \acp{PN} presented by \cite{Vinyals2015PointerNetworks} with the Transformer model presented by \citep{Vaswani2017AttentionNeed}, and are the first to demonstrate the use of Transformers to solve the \ac{CVRP}. 
The input to the encoder in the model from \cite{Kool2019AttentionProblems} is a set of nodes and the solution to the \ac{CVRP} is produced in an autoregressive manner, similar to the original Transformer paper by \cite{Vaswani2017AttentionNeed}.
Further, the authors use REINFORCE \citep{Williams1992SimpleLearning}, a policy gradient algorithm from \ac{RL}, to train the Transformer.

\citet{Kwon2020Pomo:Learning} extend the work from \citet{Kool2019AttentionProblems} and use the exact same architecture, with the two main differences being that the authors use a modified REINFORCE \citep{Williams1992SimpleLearning} algorithm and unit square transformations like rotations and reflections to introduce equivalent representations of a given graph.

The methods from \cite{Kool2019AttentionProblems} and \cite{Kwon2020Pomo:Learning} parameterize their model with Transformers and train their models with \ac{RL} and are thus referred to as \ac{DRL} methods.
In addition, they both use the \ac{PN} mechanism and attempt to solve random \ac{CVRP} instances. 
In general, their methods do not yet achieve state-of-the-art results in terms of both solution quality and speed when compared to the methods discussed in Section \ref{sec:lit_exact} and \ref{sec:lit_heur}.
For instance, \cite{Kool2019AttentionProblems} finds solutions with an average optimality gap of 3.72\% for 100-node \acp{CVRP} instances and takes 0.72 seconds on average to run on a GPU while \cite{Kwon2020Pomo:Learning} finds solutions with an average optimality gap of 0.32\% and takes 0.01 seconds on average to run on a GPU.
In comparison, \ac{HGS} obtains an average optimality gap of around 0.40\% for \ac{CVRP} instances with 100 to 330 nodes in less than three seconds \citep{Vidal2022HybridNeighborhood}.

\subsection{Decoding Strategies}
\label{sec:lit_dec}
Recent \ac{NN}-based approaches to solving routing problems model the conditional probability distribution that describes the likelihood of any given node of the problem to occur at each position of the solution sequence conditioned on the problem nodes and previously decoded solution nodes. 
As a conditional probability is modeled, sampling methods are often employed to decode a solution in an autoregressive manner. 
Adopting a sampling method naturally leads to non-deterministic solutions as we obtain a different solution trajectory for every iteration of decoding.

During decoding, the various solution trajectories generated naturally form a tree structure. 
When exploring this tree during decoding, the search space grows exponentially based on the depth of the tree, which in our context is the length of the solution sequence.
This typically renders an exhaustive tree search computationally expensive for larger problem instances.
Therefore, finding a good solution sequence with a conditional probability model requires an efficient \emph{decoding strategy} that adequately balances the trade-off between run time and quality of the sequence.

A \emph{greedy decoder} represents a very simple decoding strategy by which only the highest-probability node is considered at each position of the solution sequence.
A significant downside of this strategy is that it ignores any path dependencies during decoding.
In other words, it does not consider solution trajectories that yield better solutions by choosing low probability nodes early on in the solution sequence to make high-probability nodes accessible in subsequent parts of the solution sequence.

Another common strategy that mitigates the shortcomings of a greedy decoder is a \emph{beam search decoder} \citep{Lowerre1976TheSystem}, which is a form of modified \ac{BFS}.
In this strategy, a predefined beam size $k$ is chosen and the $k$ highest probability sequences are explored at every level of the search tree while the remaining sequences are discarded.
The main limitation of beam search is that it considers node probabilities at each level of the search tree in isolation, without any foresight into the probability distributions in subsequent levels of the tree. 
In response to this shortcoming, \cite{Lu2022NeuroLogicHeuristics} explore look-ahead heuristics that take into account future node probabilities.

Since the probability distributions of nodes at each position of the route sequence are available, one can also use a \emph{sampling decoder}, where the node realization at each position is sampled from the distribution \citep[see, e.g.,][]{Kool2019AttentionProblems}.
Extensions to sampling include \emph{top-k sampling} \citep{Fan2018HierarchicalGeneration, Holtzman2018LearningDiscriminators}, and \emph{sampling with temperature} \citep{Fan2018HierarchicalGeneration}.
In \ac{LLaMA-2}, \cite{Touvron2023LlamaModels} employ \emph{\ac{NS}} \citep{Holtzman2020TheDegeneration}, which is one of the current state-of-the-art sampling-based methods.
Since decoding is fundamentally a tree search, several authors also explore \ac{MCTS} based approaches \citep[see, e.g.,][]{Leblond2021MachineSearch, Choo2022Simulation-guidedOptimization}.

\subsection{Research Opportunities}
\label{sec:lit_gap}

In Section \ref{sec:lit_exact} and \ref{sec:lit_heur}, we discussed state-of-the-art exact and heuristic methods for solving the \ac{CVRP}. 
These methods have been extensively researched for the past few decades and involve handcrafting algorithms or local search operators.
In Section \ref{sec:lit_ML}, we reviewed state-of-the-art methods that use \ac{DL} and/or \ac{RL} to learn a policy that can approximate solutions to the \ac{CVRP}.
From our review of both streams of literature, we see two extreme paradigms emerging.
On the one hand, \ac{OR} methods have been refined with meticulous human engineering over decades to efficiently find solutions to \ac{CO} problems such as the \ac{CVRP}. 
On the other hand, \ac{ML} methods depend on minimal human engineering but demand significant computational resources.

In the era before AlexNet \citep{Krizhevsky2012ImageNetNetworks}, the canonical paper that sparked the \ac{DL} revolution, the \ac{CV} community used handcrafted methods like \ac{SIFT} \citep{Lowe2004DistinctiveKeypoints} and \ac{HOG} \citep{Dalal2005HistogramsDetection}, to achieve state-of-the-art results. 
Post AlexNet, only \ac{DL} methods have been able to achieve state-of-the-art results on current \ac{CV} problems. 

Analogously, an important research frontier is now the successful use of \ac{DL} on \ac{CO} problems. 
Developing a \ac{DL} method outperforming traditional \ac{OR} methods on a well-established \ac{CO} problem could be on the horizon.
We attempt to make a first step in this direction by filling a gap in the extant literature and applying the latest research on \acp{LLM} to the \ac{MCVRP} \citep[cf.,][]{Bengio2021MachineDhorizon}.
The \ac{MCVRP} is a special case of the \ac{CVRP} and constitutes a meaningful problem on which to focus our efforts as it has numerous impactful real-world applications and can be formulated in a manner that aligns well with \ac{LLM} architectures.


\section{Problem Definition}
\label{sec:problem}

\paragraph{The general \ac{CVRP}.}
Let $\mathcal{G} = (\mathcal{N}, \mathcal{A})$ be a symmetric undirected graph where $\mathcal{N} = \{0,...,n\}$ is the set of nodes in consideration and $\mathcal{A} = \{(i, j): i, j \in \mathcal{N}, i \neq j\}$ is the set of arcs with no self-loops connecting the nodes. 
In the generalized definition of the \ac{CVRP}, the arcs can be located anywhere in the service area, i.e., on a plane in  two-dimensional Euclidean space.
The graph is fully connected and traveling on an arc between node $i$ and node $j$ incurs symmetric costs, $c_{ij} = c_{ji}$.
Node 0 is the depot node and all other nodes are demand nodes that have an associated static demand $d_i$ and have to be visited exactly once. 
A vehicle with capacity $C$ starts at the depot, visits a sequence of demand nodes and returns to the depot under the constraint that the total demand in this route cannot exceed $C$. 
Multiple routes are executed until all demand nodes are visited and the total distance traveled is the cost of the \ac{CVRP} solution.

\paragraph{The Montreal \ac{CVRP}.}
In this paper, we consider a special experimental setup for the \ac{CVRP} referred to as the \emph{Montreal problem} in the literature \citep[see,][]{Bengio2021MachineDhorizon}.
The \ac{MCVRP} corresponds to many real-world routing problems in which the total set of potential stop locations (e.g., customer addresses) is fixed, given, and finite, while each particular instance of the problem contains only a subset of these potential stop locations with non-zero demands (e.g., the customers requiring service on a given day).

In a formal sense, the \ac{MCVRP} adheres to the structure of the \ac{CVRP}.
The primary distinction between these two problems centers on the relationship between multiple instances of the respective problem. 
Specifically, for the \ac{CVRP}, any set of problem instances operates on independent graphs.
This implies that the stop locations, which the vehicle routes are required to cover, can be distributed arbitrarily across the service region, with no inherent overlap in stops among different instances of the problem.
In contrast, for the \ac{MCVRP}, each instance of the problem represents different manifestations of non-zero demands within a subset of the same fixed and given graph $\mathcal{G'}$ of size $m'$, where $\mathcal{G'} = \left(\mathcal{N'},\mathcal{A'}\right)$. 
Consequently, each instance of the \ac{MCVRP} is confined to a subgraph $\mathcal{G}$ of size $m \ll m'$. 
This subgraph is defined as $\mathcal{G} = \left(\mathcal{N},\mathcal{A}\right)$, where $\mathcal{N} \subseteq \mathcal{N'}$ and $\mathcal{A} = \{(i, j): i, j \in \mathcal{N}, i \neq j\} \subseteq \mathcal{A'}$.

Following this formal definition of the \ac{MCVRP}, we generate problem instances by first defining a graph $\mathcal{G'}$ of size $m' = 10,001$, which consists of 10K customer nodes and 1 depot node.
Following that, we sample subgraphs $\mathcal{G}$ that range in size from $m = \{21, ..., 401\}$ (i.e., a 21 node problem instance contains 20 customer nodes and 1 depot node).
We illustrate the data generation process in greater detail in Section \ref{sec:data}.

\section{Methodology}
\label{sec:method}

In natural language, words form sentences and sentences form paragraphs.
When a human reads a piece of text, it is trivial for him or her to determine if the sentence is grammatically correct and makes sense.
It is thus natural to wonder if the recent successes in \acp{LLM} can be directly applied to the \ac{MCVRP}, where instead of training a model on a large database of natural language text, we train the model on large amounts of problem-solution pairs for the \ac{MCVRP} produced by a state-of-the-art heuristic.
We posit here that the \ac{LLM} can capture the probability distribution of solutions obtained from the heuristic and produce the order of visitation for any previously unseen set of nodes sampled from the fixed graph $\mathcal{G'}$.
In the following subsections, we describe the relevant theory and our model architecture. Specifically,
in Section \ref{sec:theory}, we detail the mathematical formulation of modeling a joint probability with Transformers.
In Section \ref{sec:arch_and_obj}, we then describe our model architecture and objective the model was trained on.
In Section \ref{sec:curric_strategy}, we elaborate on the \ac{CL} strategy, which improves efficiency in the training process and the quality of the model before outlining how solutions are obtained in Section \ref{sec:obtaining_solutions}.

\subsection{Modeling Joint Probability with Transformers}
\label{sec:theory}
Given an instance of a \ac{MCVRP}, $P$, we propose a Transformer based encoder-decoder model that aims to learn a stochastic policy to select solutions to $P$. 
We first define a feasible candidate solution $\hat{S} = (\hat{s}_1, \ldots, \hat{s}_\ell)$, where $\ell$ is the length of the solution and $\hat{s}_i \in \{0, ..., m'\}$ are the respective node IDs to be visited at position $i$.
The conditional probability of a feasible candidate solution to the \ac{MCVRP} instance, $p_\theta(\hat{S} \mid P; \theta)$, is given by
\begin{align}
    p_\theta(\hat{S} \mid P; \theta)
    &=
    \prod_{i=1}^{\ell} p_\theta(\hat{s}_i \mid \hat{s}_{i-1}, \ldots, \hat{s}_1, P; \theta),
\end{align}
where $\theta$ are the parameters of the model.
Moreover, the individual conditional probabilities $p_\theta(\hat{s}_i \mid \hat{s}_{i-1}, \ldots, \hat{s}_1, P; \theta)$ are parameterized via our Transformer-based decoder model as
\begin{align}
    P_\theta(\hat{s}_i \mid \hat{s}_{i-1}, \ldots, \hat{s}_1, P; \theta)
    &=
    H_{i, \hat{s}_i},
\end{align}
where $H_i$ is a vector that represents a finite discrete probability distribution over the node IDs in $G'$ at token position $i$ and $H_{i, \hat{s}_i}$ is a scalar that represents the probability of $\hat{s}_i$ being at token position $i$ (see Section \ref{sec:arch_and_obj}).
Our model is trained via a supervised learning procedure on a corpus of training data $(\mathcal{P}, \mathcal{S})$ consisting of pairs of \ac{CVRP} problems $P$ along with their near-optimal solutions $S^*$ and $\theta$ is optimized via \ac{SGD} using AdamW \citep{Loshchilov2019DecoupledRegularization} to minimize cross-entropy loss.

\subsection{Model Architecture and Objective}
\label{sec:arch_and_obj}

In defining our model architecture and the objective for optimizing the model, we build on the insights gathered by \citet{Raffel2020ExploringTransformer} and rely heavily on their definitions and notations.
Specifically, \cite{Raffel2020ExploringTransformer} consider two distinct objectives for unsupervised pre-training.
First, they pursue a \emph{denoising objective} \citep[see,][]{Devlin2019BERT:Understanding} for which the inputs to the model are randomly \emph{masked}, \emph{corrupted}, or left unedited.
Here, \emph{masked} means that a placeholder token that is not a word is put at the corresponding position, while \emph{corrupted} means that a random word is put at that position.
Note that the denoising objective relies on an encoder-only architecture, which contains a fully-visible mask for the attention mechanism.
Here, all tokens in the input are connected to each other.
The model then attempts to predict the correct tokens that are masked, corrupted or left unedited at these positions.
Second, they pursue an \emph{\ac{LM} objective} analogous to what we described in Section \ref{sec:theory}.
As this models a conditional probability, a causal mask is used in the attention mechanism, such that a token at any given position can only view previous tokens and not future tokens.
Through extensive experiments, \cite{Raffel2020ExploringTransformer} conclude that the combination of an encoder-decoder architecture with a denoising objective yields the highest performance on a set of benchmark \ac{NLP} tasks \citep[cf. Table 2 in ][]{Raffel2020ExploringTransformer}.
Given the generally superior performance exhibited by an encoder-decoder architecture in their analyses, we also adopt this architecture in our work.
However, since a denoising objective is not suitable for our problem structure, we rely on an \ac{LM} objective.
We further elaborate on our architecture and the objective in the next paragraphs.

\paragraph{Encoder-decoder architecture.}
Following the insights from \cite{Raffel2020ExploringTransformer}, we adopt the encoder-decoder architecture.
We first introduce two concepts at a high level that are necessary to understand the encoder-decoder architecture: the embedding layer and the output layer.
In \ac{NLP}, words are represented as a one-hot vector and \ac{ML} models typically operate on feature vectors that contain continuous values.
Therefore, it is common to have a learned matrix of size $\mathbb{R}^{|G'| \times D}$ to transform the features into vectors with dimension $D$, where $|G'|$ is the cardinality of the fixed graph described in Section \ref{sec:problem}.
The encoder-decoder then processes the vectors of size $D$ through multiple encoder and decoder blocks and finally outputs embeddings of size $D$.
To translate this representation back into the original vocabulary space, a learned matrix of size $\mathbb{R}^{D \times |G'|}$ transforms these embeddings and a softmax can be applied to obtain word predictions.
The learned matrix of size $\mathbb{R}^{|G'| \times D}$ is commonly referred to as an embedding layer and the learned matrix of size $\mathbb{R}^{D \times |G'|}$ is commonly referred to as the output layer.
In practice, these matrices share weights and one is simply the transpose of the other.

In this work, we make two slight modifications to the encoder-decoder architecture as our problem is not an \ac{NLP} problem.
First, as the inputs to our problem are already feature vectors with continuous values, and not one-hot vectors representing a vocabulary, the first embedding layer is unnecessary.
Second, the embeddings from the encoder and decoder are both passed through the output layer for prediction of the corresponding node IDs.
Therefore, the output layer is shared by both the encoder and the decoder.

\begin{figure}[htbp]
    \centering
    \includegraphics[width=\textwidth]{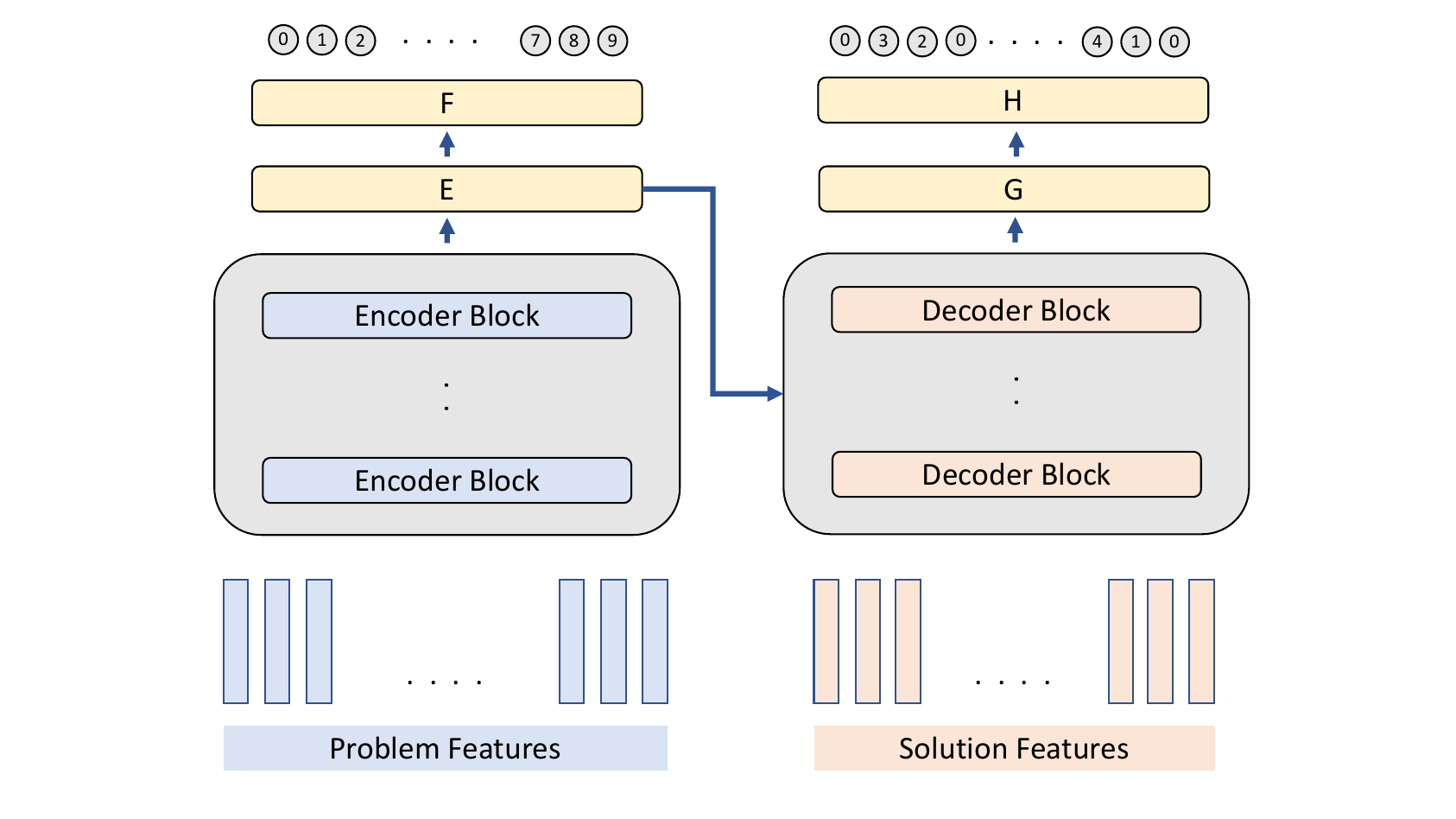}
    \caption{Encoder-Decoder Transformer Architecture for the \ac{CVRP}.}
    \label{fig:architecture}
\end{figure}

Figure \ref{fig:architecture} provides an illustrative summary of our overall methodological approach.

Following the terminology used in \cite{Vaswani2017AttentionNeed} and \cite{Raffel2020ExploringTransformer}, the encoder consists of a series of encoder blocks, which contain fully-visible \ac{MHA} and feed-forward layers.
The decoder consists of a series of decoder blocks, which contain causal \ac{MHA}, fully-visible \ac{MHA} and feed-forward layers.
In addition, the matrices $E$, $F$, $G$ and $H$ in Figure \ref{fig:architecture} are given by
\begin{align*}
    F &= \mathrm{softmax}(E \cdot W_o), & E \in \mathbb{R}^{m \times D}, W_o \in \mathbb{R}^{D \times |\mathcal{G'}|},\\
    H &= \mathrm{softmax}(G \cdot W_o + M), & G \in \mathbb{R}^{n \times D}, W_o \in \mathbb{R}^{D \times |\mathcal{G'}|}, M \in \mathbb{R}^{D \times |\mathcal{G'}|}
\end{align*}
where $E$ represents the node embeddings output by the encoder after transforming the problem features via a series of encoder blocks;
$G$ represents the node embeddings output by the decoder after transforming the solution features via a series of decoder blocks, with $E$ being used as the keys and values in the attention mechanism;
$W_o$ represents the shared output layer;
$m$ and $n$ represent the total number of nodes in the problem and solution respectively, where $n > m$ as the solution involves returning to the depot;
$M$ represents a mask that is applied to prevent nodes that are not in the problem and infeasible nodes from being predicted.

Observe that $E$ is used in the attention mechanism in the decoder, which directly follows the original Transformer paper \citep[see,][for more details]{Vaswani2017AttentionNeed}.
Further, observe how $E \cdot W_o$ and $G \cdot W_o$ are matrices of size $\mathbb{R}^{m \times |\mathcal{G'}|}$ and $\mathbb{R}^{n \times |\mathcal{G'}|}$, respectively, and the softmax is applied to the resulting matrices across the dimension corresponding to the size of $|G'|$  (i.e. the matrix of size $\mathbb{R}^{m \times |\mathcal{G'}|}$ has $m$ rows that each sum to one).
In particular, each vector in $F$ and $H$ represents a finite discrete probability distribution over the node IDs in $G'$ at a given token position.

\paragraph{Objective.}
For the encoder, we are unable to use a denoising or an \ac{LM} objective as the encoder takes as input the problem tokens of a subgraph $\mathcal{G'}$.
It is easy to see why this is the case as applying a mask or simply replacing a token with another random token would result in an entirely different problem.
Similarly, using an \ac{LM} objective would impose a structure on the subgraph $\mathcal{G'}$ such that nodes do not have full visibility of each other.
For the decoder, we follow \cite{Kool2019AttentionProblems} and \cite{Kwon2020Pomo:Learning} and chose to define a model with an \ac{LM} objective and thus are unable to use the denoising objective.
Future extensions can model the decoder differently and enable the use of a denoising objective. 
In addition, inspired by \ac{CV} techniques, we add data augmentations by rotating the entire subgraph $\mathcal{G'}$ by a random angle to enable the model to be invariant to rotations.

Finally, as we are predicting node IDs on both the encoder and decoder, we use the cross-entropy loss for a single token,
\begin{align}
\label{eqn:cross_entropy}
L(p, q) = -\sum_{n=0}^{|\mathcal{G'}|} p(n) \log(q(n)),
\end{align}
where $L$ represents the cross-entropy between the true distribution $p$ and the predicted distribution $q$, $|\mathcal{G'}|$ represents the cardinality of the subgraph $\mathcal{G'}$, and $n$ represents the respective node IDs.
In scenarios with one-hot encoded labels, $p(n) = 1$ for the correct node and $0$ for all other nodes. 
$q(n)$ represents the predicted probability of the token belonging to node $n$, which is derived from the softmax output of the model.
The cross-entropy loss is computed and the model is optimized via \ac{SGD} with the AdamW optimizer \citep{Loshchilov2019DecoupledRegularization}.

\paragraph{Inputs and outputs.}
The inputs to the encoder are the problem features and the inputs to the decoder are the solution features.
We characterize each token (i.e., each node) with 9 features for consistency.
These node features are $(x, y, d, t, \kappa, \gamma, \omega, c, a)$. 
$x$ and $y$ represent the respective coordinates of a node. Their values are normalized into a range of $[0,1]$.
$d$ represents the demand of a node and is normalized into a range of $[0,1]$, where a value of 1 corresponds to the maximum capacity of the vehicle.
$t$ is a binary variable that represents the type of the node, with a value of 1 for the depot and 0 otherwise.
Future extensions of this work could include multi-depot scenarios, where each depot is represented by a one-hot vector.
$\kappa$ represents the normalized distance of a node to the depot.
$\gamma$ and $\omega$ respectively represent the cosine and sine angle of a node with respect to the depot.
$c$ represents the capacity utilization of the vehicle at any point in a particular solution and is in the range of $[0,1]$, where a value of 1 represents that the vehicle is full.
$a$ represents the total normalized demand fulfilled in the entire network so far.
Problem tokens can be fully described by $(x, y, d, t, \kappa, \gamma, \omega)$ while the other features are set to zero.
Solution tokens are fully described by all 9 features.

\subsection{\acl{CL} Strategy}
\label{sec:curric_strategy}
In this work, we leverage the \ac{CL} strategy first proposed by \citet{Bengio2009CurriculumLearning}.
\ac{CL} assumes that when trying to learn a task, humans and animals learn better when the examples are not presented randomly but in a certain order.
Analogously, we are training an \ac{ML} model by presenting it with training data in a predefined order.
The \ac{CVRP} lends itself to a natural ordering of the problem instances, where smaller problem instances are easier to solve than larger ones.
There are two advantages in adopting the \ac{CL} strategy.
First, \citet{Bengio2009CurriculumLearning} hypothesize that \ac{CL} enables faster convergence as well as an increase in the quality of minima obtained.
Second, our model accepts inputs and outputs of varying sizes in batch training, which is achieved through padding.
As such, the difference between the smallest and largest training sample determines the amount of padding required and this can cause a lot of unnecessary computation.
With \ac{CL}, we are able to reduce the amount of unnecessary computation as training samples with a smaller size have less padding.
Further, \ac{CL} enables us to use a larger batch size in early stages of training as the savings from additional padding permit more training samples in a single batch, allowing the model to potentially see more training samples within the same computational budget.

\subsection{Obtaining Solutions}
\label{sec:obtaining_solutions}
In Section \ref{sec:theory}, we describe how our method models the conditional probability of the next node to visit at a given token position, conditioned on the current partial solution tokens and all the problem tokens.
To obtain solutions to new problem instances based on our trained model, we follow the existing literature which distinguishes two general approaches.
First, we can employ a simple \emph{\ac{GS}} approach by taking the $\arg\max$ of the conditional probability vector obtained at each token position when decoding autoregressively.
This approach is deterministic and computationally efficient as only a single sample of the trajectory is required.
Second, we can follow a state-of-the-art sampling technique from \ac{NLP} and employ \emph{\ac{NS}} \citep{Holtzman2019TheDegeneration}, also known as \emph{top-p sampling}, which first truncates the probability vector up to a threshold $p$, normalizes this distribution and then samples from it.
This approach is non-deterministic and incurs a significantly higher computational cost, which depends on the number of trajectory samples.
In return, a higher number of samples typically leads to better performance in terms of the quality of the solutions obtained.
In addition, when sampling multiple solutions to a given problem instance, we also randomly rotate the graph to increase variability in the sampling process (i.e., some angles for a given graph may yield better solutions).
As mentioned in Section \ref{sec:lit_dec}, other potential decoding strategies include \emph{beam search} \citep{Lowerre1976TheSystem}, \emph{top-k sampling} \citep{Fan2018HierarchicalGeneration, Holtzman2018LearningDiscriminators} and \emph{sampling with temperature} \citep{Fan2018HierarchicalGeneration}.
In this work, we do not pursue these strategies and choose to follow \ac{LLaMA-2}, a state-of-the-art \ac{LLM} that uses \emph{\ac{NS}}. 
We also include \emph{\ac{GS}} as a simple baseline to assess the performance of the model.


\section{Computational Experiments}
\label{sec:experiment_setup}
In the following, we describe the large-scale experiments we conducted to demonstrate the performance and potential of our method in utilizing an \ac{LLM} to generate high-quality solutions for \ac{MCVRP} instances.
In Section \ref{sec:data}, we discuss our data preparation method.
In Section \ref{sec:params}, we discuss the model and training parameters used.
Finally, in Section \ref{sec:benchmarks}, we discuss the benchmarks utilized for comparing the performance of \ac{FM-MCVRP}.

\subsection{Generating Data}
\label{sec:data}
We primarily follow \citet{Nazari2018ReinforcementProblem} in generating problem instances, a benchmark also adopted by \citet{Kool2019AttentionProblems} and \citet{Kwon2020Pomo:Learning}, which are considered canonical works for \ac{DRL} on the \ac{CVRP}.
However, recall that \citet{Kool2019AttentionProblems} and \citet{Kwon2020Pomo:Learning} train a distinct model for each problem size and fix the capacity.
In contrast, our research contribution involves training a single model across various problem sizes ranging from 20 to 400 customer nodes and varying vehicle capacities.
We choose this approach to mimic real-world delivery operations, which typically involve routes of varying lengths (e.g., due to varying customer density and drop sizes) and heterogeneous vehicle capacities (e.g., due to mixed fleets). In such a context, it is essential to be able to train and use a single unified model for the entire service area of a given city.

\paragraph{Problem instances.}
We first generate a graph $\mathcal{G'}$ of size 10K (not including the depot).
The locations of the nodes in $\mathcal{G'}$ fall within a unit square, with the depot node being placed in the middle of the unit square. 
The locations of the customer nodes are randomly sampled from a uniform distribution over the unit square.
Subsequently, a problem instance is formed by including the depot and sampling without replacement a set of customer nodes from $\mathcal{G'}$.
The demand for each customer node is uniformly sampled from the set $\{1, ..., 9\}$.
Note that our work deviates from previous works in the literature in the way the vehicle capacity is defined for a given problem instance. 
Unlike \cite{Nazari2018ReinforcementProblem}, \cite{Kool2019AttentionProblems} and \cite{Kwon2020Pomo:Learning}, who associate a specific vehicle capacity with each problem size, we choose capacity $C$ for each problem instance by uniformly sampling within the ranges defined for any problem instance size $n$ in Table \ref{table:node_capacity}.

\begin{table}[htbp]
\centering
\begin{tabularx}{0.6\textwidth}{XX}
\toprule
{\bf Number of Nodes ($n$)} & {\bf Capacity ($C$)}\\
\midrule
$20 \leq n < 50$ & $30 \leq C < 40$\\
$50 \leq n < 100$ & $40 \leq C < 50$\\
$100 \leq n < 200$ & $50 \leq C < 60$\\
$200 \leq n \leq 400$ & $60 \leq C < 70$\\
$401 \leq n \leq 1,000$ & $70 \leq C < 80$\\
\bottomrule
\end{tabularx}
\caption{The range of capacities corresponding to the number of nodes.}
\label{table:node_capacity}
\end{table}

\paragraph{Problem-solution pairs.}
Following the definition of the \ac{MCVRP} (see Section \ref{sec:problem}), we generate 100K unique problem instances on $\mathcal{G'}$ for every possible problem size in the range $\{20, ..., 400\}$ customers, leading to a total of 38.1M instances.
For each problem instance, we obtain a solution using \ac{HGS} \citep{Vidal2022HybridNeighborhood} with a time limit of 5 seconds.
We chose \ac{HGS} over \ac{LKH-3} for our solution generation for two reasons.
First, \ac{HGS} is an algorithm with generally better solution quality over \ac{LKH-3} given a fixed time limit, as shown in \citet{Vidal2022HybridNeighborhood}.
Second, \ac{HGS} enables us to control the time the algorithm is allowed to run for, whereas the run time in \ac{LKH-3} can only be controlled indirectly through the number of trials and runs.

We define $T_i$ as the dataset containing all problem-solution pairs of size $i$.
Further, we define $T^{\text{trunc}}_i \subset T_i$ as the dataset containing a random subset of 1K problem-solution pairs of size $i$.
$T^{\text{trunc}}_i$ is necessary for encoder pre-training (see Section \ref{sec:arch_and_obj}) as we want the model to train on a large variety of problem sizes within a reasonable time.

The decision for a 5 second time limit being imposed on \ac{HGS} is a hyperparameter that can be tuned.
We base our choice of this parameter value on two factors: computational cost and solution quality.
We follow \citet{Vidal2022HybridNeighborhood} and measure solution quality by the percentage gap of the solution compared to the \ac{BKS}. This percentage gap is given by $\text{Gap} = 100 \times (z - z_{\text{BKS}})/z_{\text{BKS}}$, where $z$ is the solution value of the algorithm and $z_{\text{BKS}}$ is the \ac{BKS} value for this problem instance.
For data generation, we rely on 48 parallel processes on Intel Xeon Platinum 8260 processors (48 CPU cores) on a total of 8 nodes, giving us a total of 384 parallel processes during data generation.
From the perspective of computational cost, generating solutions for 38.1M instances on the aforementioned compute infrastructure under a 5 second time limit per instance results in a total run time of slightly more than one day.
From the perspective of solution quality, we did not want to generate solutions that are too close to optimality or the \ac{BKS}.
This is by design as we want to show that our proposed method can learn from a large dataset of good, but sub-optimal and hence relatively inexpensive solutions and outperform the quality of the solutions it has been trained on.

\subsection{Model and Training Parameters}
\label{sec:params}

\paragraph{Model parameters.}
Both encoder and decoder in our model use \ac{MHA} with 12 attention heads, 12 layers, an embedding dimension of 768, a feed-forward layer dimension of 3,072, a dropout probability of $0.1$, and the final layer having a dimension of 10,001, representing the 10K potential customer nodes and the depot.
With these configurations, our model has 206M parameters in total.

\paragraph{Training parameters.}
All our models are trained on 16 Tesla V100-PCIE-32GB GPUs on 
\ac{MITSC} \citep{Reuther2018InteractiveAnalysis}.
As 
\ac{MITSC} 
has a strict time limit of 96 hours for a job, we designed our training process with these constraints in mind.
We leverage the \ac{CL} strategy described in Section \ref{sec:curric_strategy} and define a curriculum as
\begin{align*}
    Cr_i &= \bigcup_{j=t}^{i} T_j, \quad Cr^{\text{trunc}}_i = \bigcup_{j=t}^{i} T^{\text{trunc}}_j, & 20 \leq i \leq 400, 
\end{align*}
where $Cr_i$ and $Cr^{\text{trunc}}_i$ contain 100K and 1K problem-solutions pairs for each size from size 20 to size $i$ (inclusive) customer nodes, respectively.

Our training parameters closely follow \citet{Raffel2020ExploringTransformer} and are summarized in Table \ref{tab:training_params}.
At a high level, training can be broken down into two large phases: encoder pre-training (Phase I) and encoder-decoder finetuning (Phase II-A through II-C).
As shown in Table \ref{tab:training_params}, all parameters except the curriculum are essentially the same across Phase II.
The proposed split is due to the 96 hour limitation on 
\ac{MITSC}.
\citet{Raffel2020ExploringTransformer} use a batch size of 128, whereas we use an effective batch size of 256 (batch size of 16 per GPU and a total of 16 GPUs), giving a learning rate scaling factor of $\sqrt{2}$ (see Appendix \ref{app:choosing_hyperparams}).
We opted to scale the learning rate only in the finetuning phase as the pre-training learning rates were sufficiently high in encoder pre-training.
During pre-training, the learning rate follows the \ac{T5} schedule of $\frac{1}{\sqrt{\max(n, k)}}$, where $n$ is the current training iteration and $k$ is the number of warm-up steps.
As we use 10K warm-up steps \citep{Raffel2020ExploringTransformer}, this means that the learning rate is kept constant at 0.01 for the first 10K warm-up steps and thereafter decays exponentially to 0.002.
During finetuning, the learning rate is kept constant at $\sqrt{2}\times10^{-3}$, which follows \ac{T5} as well but with an additional scaling factor.
We also use the AdamW \citep{Loshchilov2019DecoupledRegularization} optimizer and clip gradients with a norm larger than $1.0$.
The optimizer, learning rate schedules and batch sizes were carefully chosen based on existing literature and we refer interested readers to Appendix \ref{app:choosing_hyperparams} for details.

\begin{table}[htbp]
    \centering
    \footnotesize
    \begin{tabularx}{\textwidth}{ccccccccX}
        \toprule
        {\bf Phase} & {\bf Curriculum} & {\bf Model} & {\bf BSZ/GPU} & {\bf Peak LR} & {\bf Min LR} & {\bf Warm-up} & {\bf Rotation} & {\bf Time} \\
        \midrule
        I & \([Cr^{\text{trunc}}_{20}, \ldots, Cr^{\text{trunc}}_{400}]\) & Enc & 16 & 0.01 & 0.002 & 10K & No & 52hr \\
        II-A & \([Cr_{20}, \ldots, Cr_{50}]\) & Enc-Dec & 16 & $\sqrt{2}\times10^{-3}$ & $\sqrt{2}\times10^{-3}$ & 0 &  Yes & 59hr \\
        II-B & \(Cr_{200}\) & Enc-Dec & 16 & $\sqrt{2}\times10^{-3}$ & $\sqrt{2}\times10^{-3}$ & 0 & Yes & 26hr \\
        II-C & \(Cr_{400}\) & Enc-Dec & 16 & $\sqrt{2}\times10^{-3}$ & $\sqrt{2}\times10^{-3}$ & 0 & Yes & 96hr \\
        \bottomrule
    \end{tabularx}
    \caption{Training parameters used.}
    \label{tab:training_params}
\end{table}

\subsection{Performance Benchmarks}
\label{sec:benchmarks}
Throughout our analysis, we compare the performance of our proposed \ac{FM-MCVRP} against two state-of-the-art heuristics, \ac{HGS} and \ac{LKH-3}, and the method presented by \cite{Kool2019AttentionProblems}, which is a recent \ac{DRL} approach and referred to as \ac{AM} in the following.
Note that the publicly available weights for \ac{AM} are trained for the general \ac{CVRP}. 
Therefore, we need to retrain \ac{AM} for the \ac{MCVRP}.
Since the model in \ac{AM} caters to a unique problem instance size, we in fact retrain a separate \ac{AM} for every instance size considered in our analysis.
Further details on the retraining process can be found in Appendix \ref{app:retraining-am}.

For each of these benchmark methods, we discuss model performance for instance sizes of 20, 50, 100, 200, 400, 600 and 800 customer nodes.
To assess whether our observed model performance is systematic rather than just a coincidental artifact of the specific problem instances we are solving, we solve 1,000 problem instances for each instance size of 20, 50, 100, 200, and 400 customers, respectively. 
To avoid excessively large computation times, we reduce the number of instances solved to 100 for the larger problem instances of 600 and 800 customers.

Conditional probability models like our \ac{FM-MCVRP} and \ac{AM} generally utilize sampling methods to obtain the best results (see Section \ref{sec:obtaining_solutions}).
In this work, we use \ac{NS} \citep{Holtzman2020TheDegeneration} in \ac{FM-MCVRP} and the default sampling method in \ac{AM}, which samples from the conditional probability vector without modifications.
While they do not rely on sampling per se, both \ac{HGS} and \ac{LKH-3} use random seeds when finding an initial solution.
Thus, the results from all of the methods we seek to compare are non-deterministic.
Therefore, to ensure an equitable comparison of their performance, we generate 1, 100, and 1,000 solutions for any given problem instance with each of these methods, respectively.
When discussing our results, we report the best found solution in terms of solution value for the currently discussed sample size.

Given that both \ac{HGS} and \ac{LKH-3} are improvement methods, the quality of the solutions they find predominantly depends on the amount of run time they are granted before terminating the solution process. 
Since our experimental setup is built on the premise that \ac{FM-MCVRP} is trained in a supervised manner on sub-optimal solutions, we choose to obtain these solutions by imposing a tight computational budget of 5 seconds on \ac{HGS}.

For all of the \ac{LKH-3} results discussed in our analysis, we use the default configuration of the algorithm, as proposed by \cite{Helsgaun2017AnProblems}, which sets the maximum number of trials to the number of customer nodes in the problem instance.

\section{Results and Discussion}
\label{sec:results}
In the following, we present and discuss some of the most relevant results from our computational experiments.
We structure our findings loosely according to the main contributions of our work, as stated in Section \ref{sec:introduction}.

In Section \ref{sec:double_descent}, we discuss the double descent phenomenon that can be observed in our model convergence and its importance in avoiding premature termination of the training process for these types of models.
In Section \ref{sec:outperform_heuristic}, we show that \ac{FM-MCVRP} is able to produce solutions that are of higher quality than the solutions it was trained on.
In Section \ref{sec:distance_distribution}, we further compare the distance distribution of solutions obtained with \ac{FM-MCVRP} and \ac{HGS} to demonstrate that \ac{FM-MCVRP} is competitive with \ac{HGS} even under a less restrictive computational budget.
In Section \ref{sec:generalizing_to_larger_problems}, we show that \ac{FM-MCVRP} can generalize well to larger problem instances.
In Section \ref{sec:extended_comparison}, we extend our comparisons to \ac{LKH-3} and \ac{AM} and show that \ac{FM-MCVRP} is still competitive with recent state-of-the-art methods. 

\subsection{Convergence and Double Descent}
\label{sec:double_descent}

\begin{figure}[htbp]
    \centering
    \begin{subfigure}{0.5\textwidth}
        \centering
        \includegraphics[width=\textwidth]{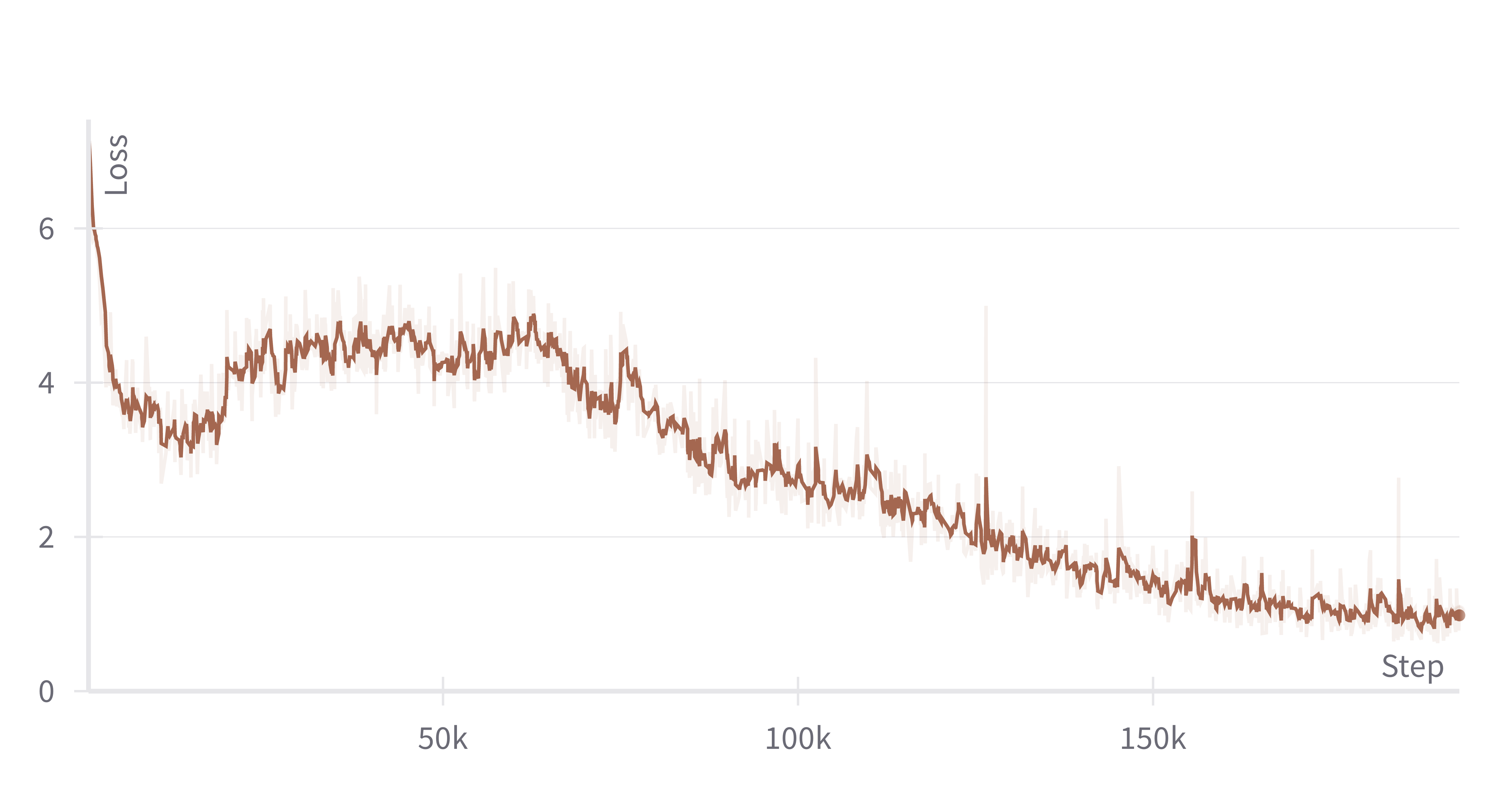}
        \captionsetup{font=scriptsize}
        \subcaption{Phase II-A Problem Tokens}
        \label{fig:p2a-prob}
    \end{subfigure}%
    \hfill
    \begin{subfigure}{0.5\textwidth}
        \centering
        \includegraphics[width=\textwidth]{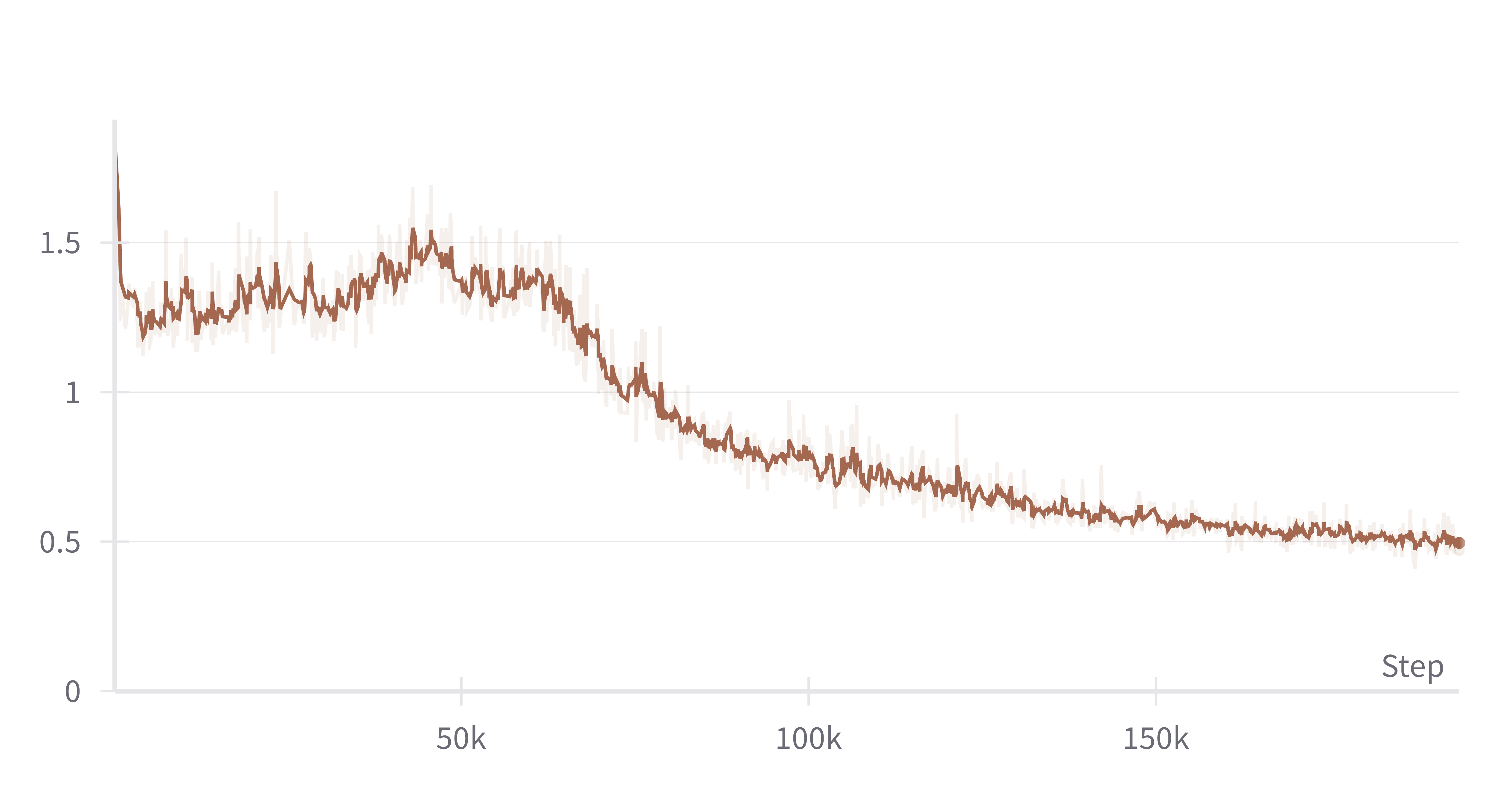}
        \captionsetup{font=scriptsize}
        \subcaption{Phase II-A Solution Tokens}
        \label{fig:p2a-sol}
    \end{subfigure}%
    \\
    \begin{subfigure}{0.5\textwidth}
        \centering
        \includegraphics[width=\textwidth]{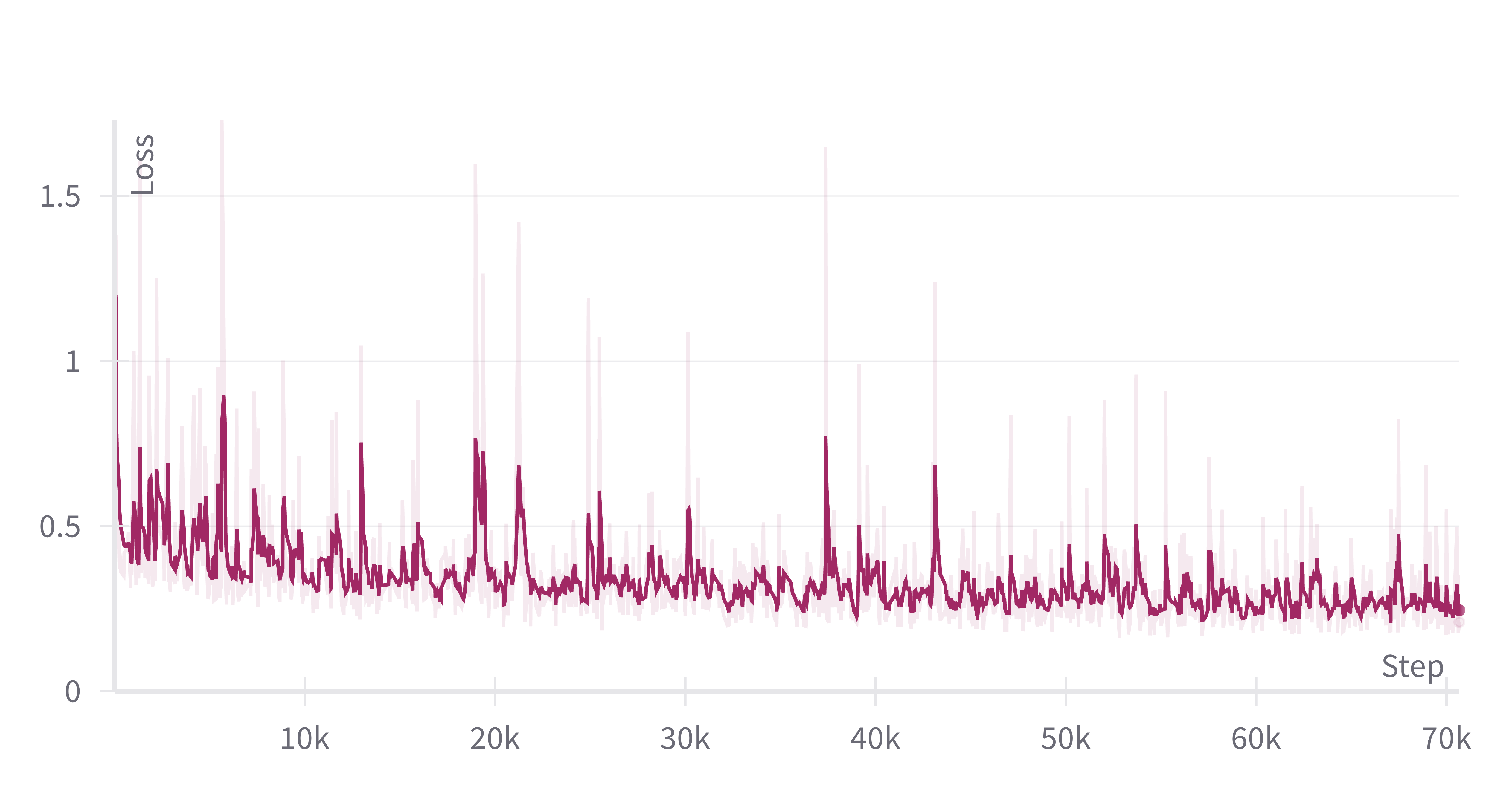}
        \captionsetup{font=scriptsize}
        \subcaption{Phase II-B Problem Tokens}
        \label{fig:p2b-prob}
    \end{subfigure}%
    \hfill
    \begin{subfigure}{0.5\textwidth}
        \centering
        \includegraphics[width=\textwidth]{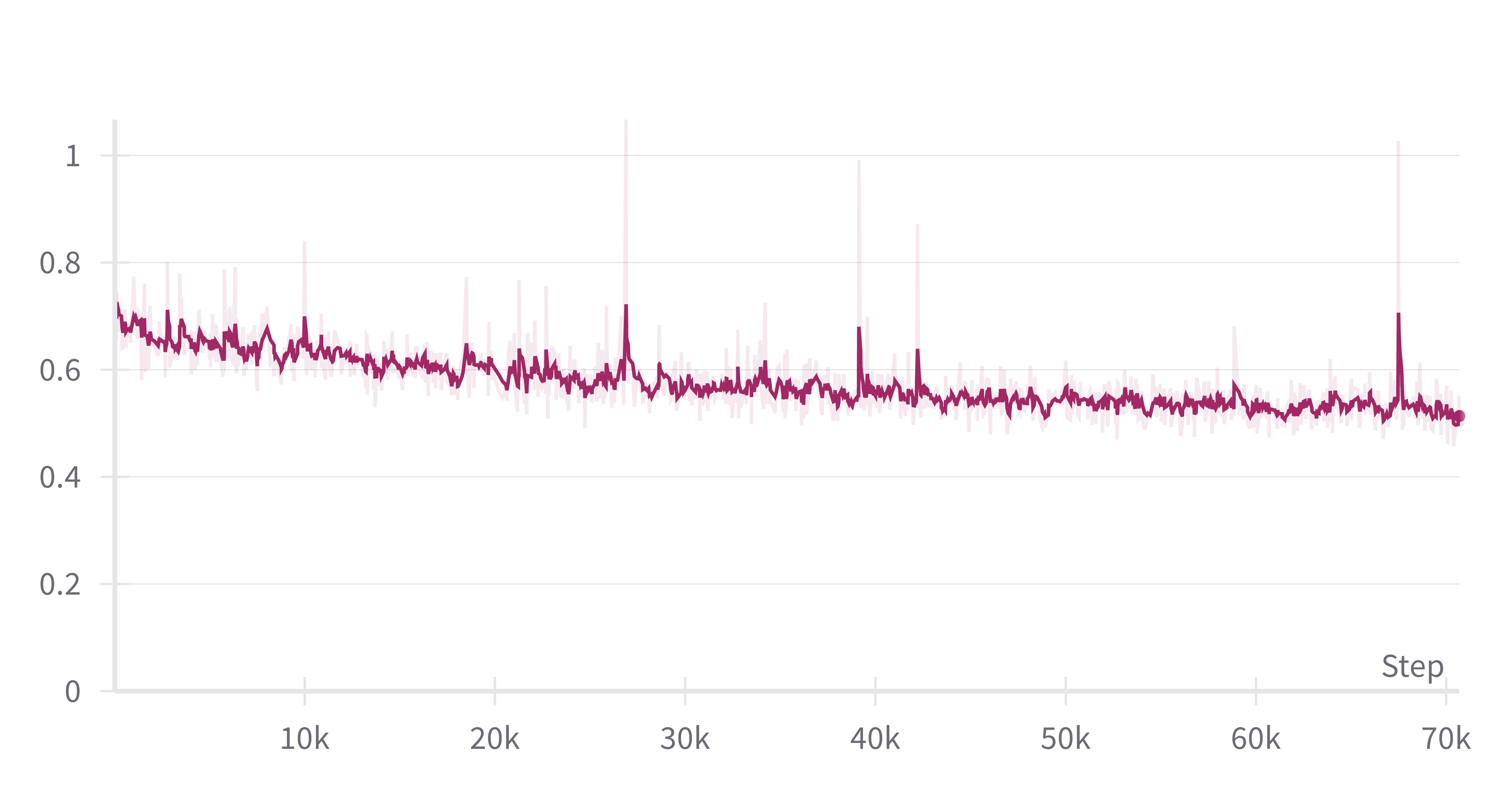}
        \captionsetup{font=scriptsize}
        \subcaption{Phase II-B Solution Tokens}
        \label{fig:p2b-sol}
    \end{subfigure}%
    \\
    \begin{subfigure}{0.5\textwidth}
        \centering
        \includegraphics[width=\textwidth]{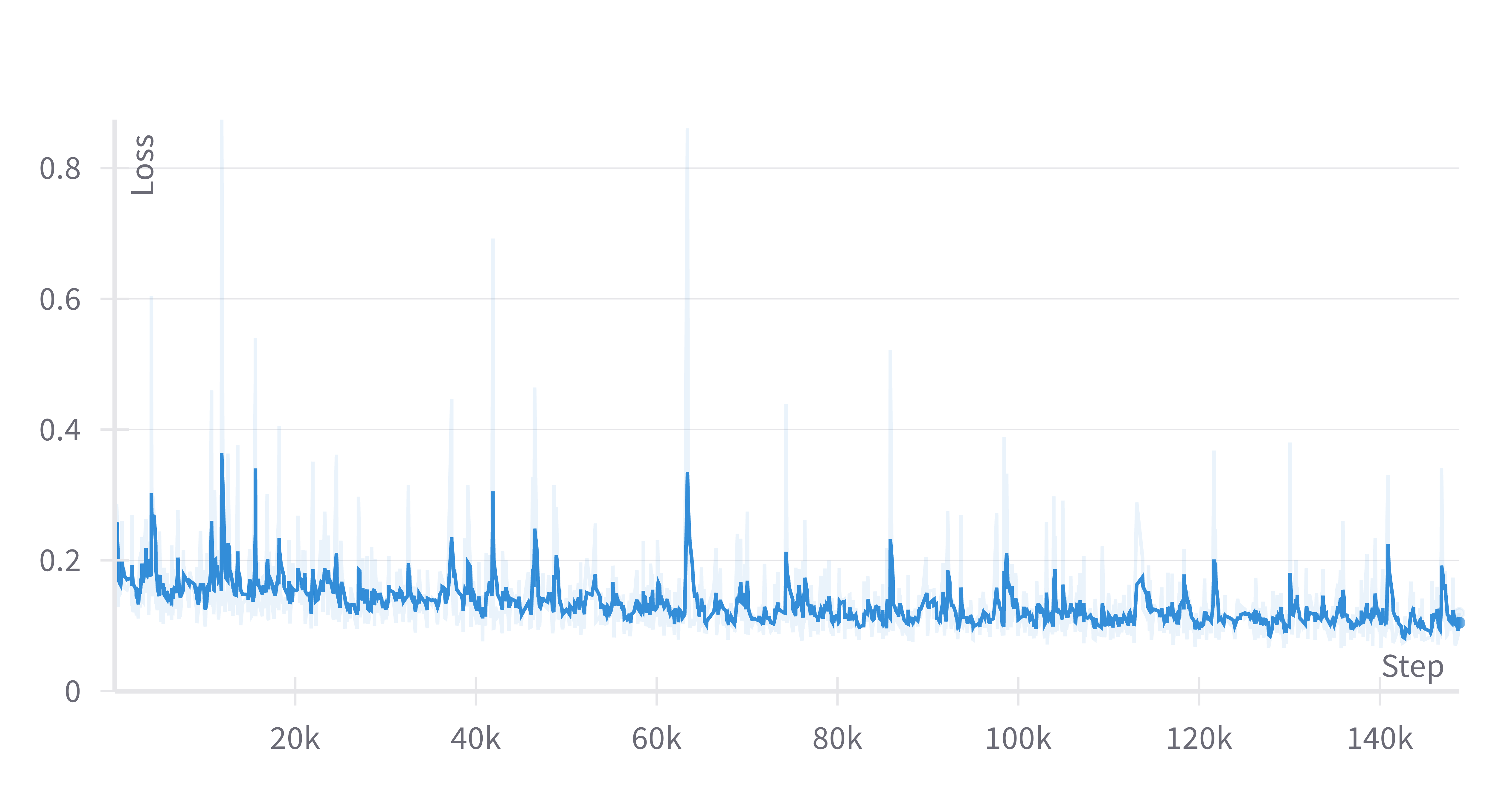}
        \captionsetup{font=scriptsize}
        \subcaption{Phase II-C Problem Tokens}
        \label{fig:p2c-prob}
    \end{subfigure}%
    \hfill
    \begin{subfigure}{0.5\textwidth}
        \centering
        \includegraphics[width=\textwidth]{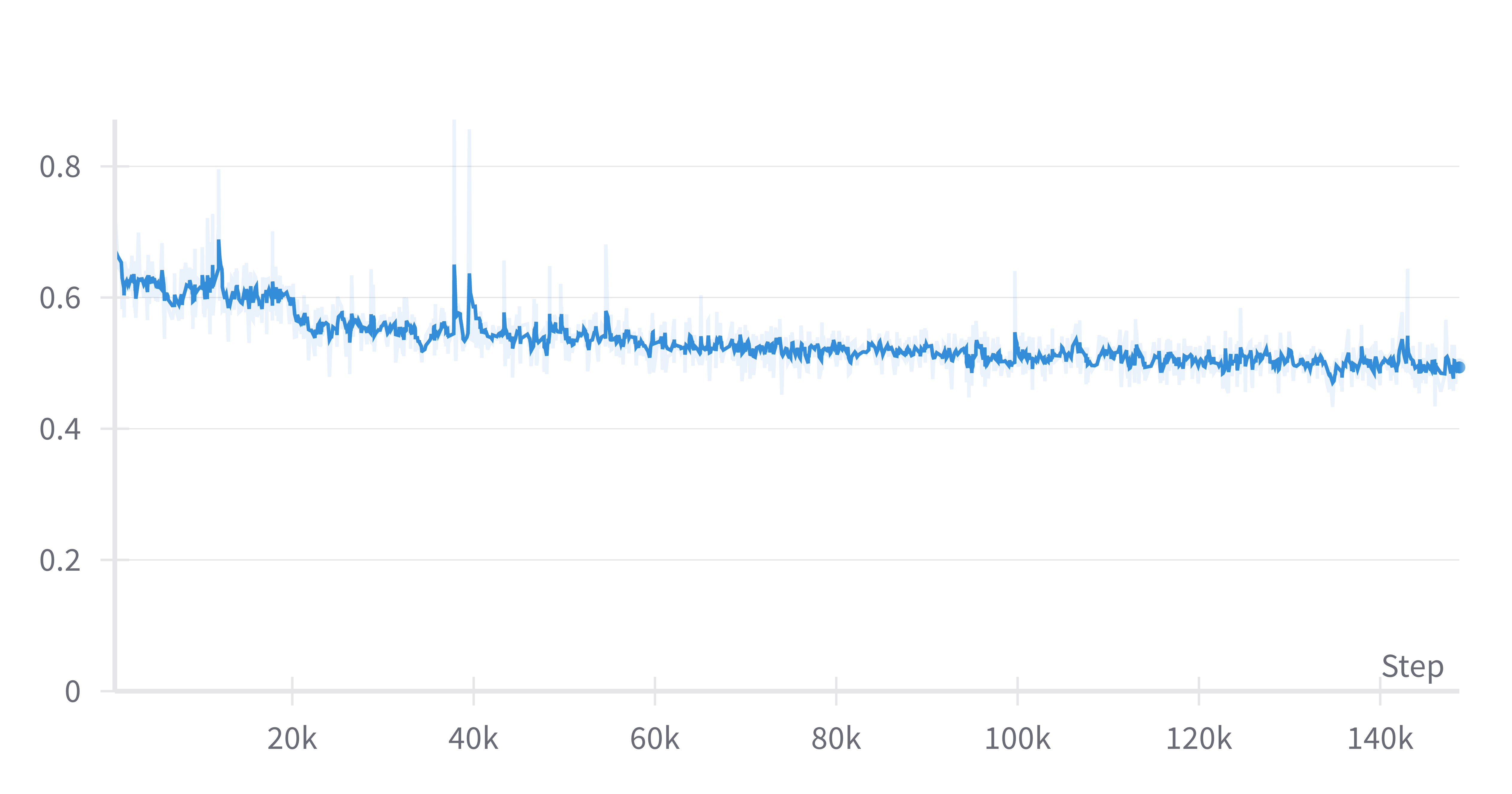}
        \captionsetup{font=scriptsize}
        \subcaption{Phase II-C Solution Tokens}
        \label{fig:p2c-sol}
    \end{subfigure}%
    \caption{Training loss for problem (left) and solution (right) tokens in Phase II.}
    \label{fig:grid}
\end{figure}

A first key insight from our numerical analysis pertains to the existence of a pronounced double-descent behavior in the convergence of the training loss of \ac{FM-MCVRP}.
Following \citet{Raffel2020ExploringTransformer} and \citet{Touvron2023LlamaModels}, Figure \ref{fig:grid} shows the training loss obtained for the various training phases.
Specifically, we highlight the loss curves in Phase II-A (Figures \ref{fig:p2a-prob} and \ref{fig:p2a-sol}) as we clearly observe the double descent phenomenon discussed by \cite{Nakkiran2021DeepHurt}.
This is an important insight to bear in mind when training large models as the convergence of these models can sometimes appear to plateau and thus trigger an early termination of model training when it is in fact going through double descent.
In the context of our proposed model, for Phase II-A, after 20 hours of training, the loss appears to be plateauing and training might be terminated if one was not aware of the double descent phenomenon.

Observe in our training process that each successive phase in Phase II introduces new problem-solution pairs.
For example, Phase II-C contains problem-solution pairs of size 201 to 400 customer nodes for the first time.
Therefore, the training loss in Phase II-C (Figures \ref{fig:p2c-prob} and \ref{fig:p2c-sol}) is a rough approximation of the performance of the model as more than half of the samples have only been seen once.
To interpret the loss values, we remind the reader of the definition of cross-entropy loss we use in Equation \eqref{eqn:cross_entropy}.
For problem tokens, the final loss converges to a value of approximately 0.09, which translates to identifying the correct node ID with a probability of 0.91 on average.
For solution tokens, the final loss converges to a value of approximately 0.50, which translates to identifying each token in the \ac{HGS} solution with a probability of 0.61 on average.
This implies that the model is fairly confident of choosing a \ac{HGS} solution from the training data but also has a relatively large margin of 0.39 on average to explore other nodes.
An interesting extension of this work could include training the model with significantly more data and over a longer period of time and analyze the value to which the loss converges to.

\subsection{Solution Quality Under a Tight Computational Budget}
\label{sec:outperform_heuristic}

A second key result from our numerical analysis is that once trained, our proposed model can outperform the solutions it was trained on.
As described in Section \ref{sec:data}, we generate our training data by obtaining a single solution per problem instance with \ac{HGS} under a time limit of 5 seconds for two reasons. 
First, it is computationally efficient to obtain \ac{HGS} solutions under this computational budget and the solutions obtained are generally of good quality but typically sub-optimal.
Second, we also purposefully aimed for this good yet sub-optimal solution quality in our training data to emulate the characteristics of many real-world routing datasets available to companies in practice.
In Table \ref{table:key_result}, we compare the solutions obtained by \ac{FM-MCVRP} with the sub-optimal solutions it was previously trained on. 
Here, we make three important observations.

First, similar to what \citet{Kool2019AttentionProblems} and \citet{Kwon2020Pomo:Learning} observe, 
the gap in solution quality between the solutions found by \ac{FM-MCVRP} and the solutions it was trained on decreases as the instance size gets larger.
This implies that under a tight computational budget, the solution quality of a state-of-the-art heuristic such as \ac{HGS} declines in instance size at a faster rate than the quality of the solutions found with our method.

\begin{figure}[htbp]
    \centering
    \includegraphics[width=0.75\textwidth]{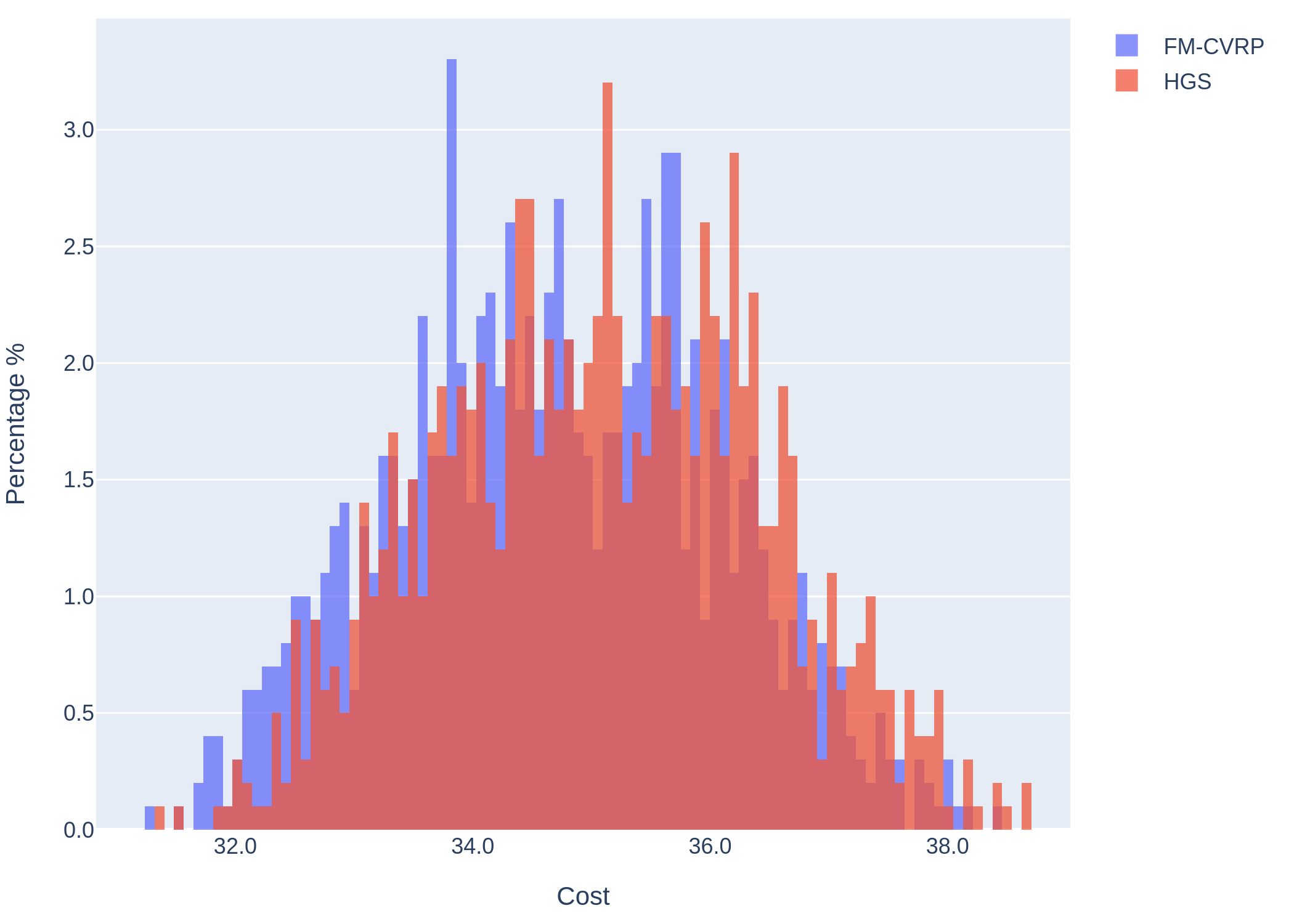}
    \caption{Distance distributions of \ac{FM-MCVRP} (\ac{NS}, 1,000 samples) and \ac{HGS} (single run) on 1,000 400-customer problem instances.}
    \label{fig:hgs_1_vrpt5_1000}
\end{figure}

Second, under an \ac{NS} decoding strategy with $100$ or $1,000$ samples, \ac{FM-MCVRP} is shown to outperform the solutions it was trained on for large problem instances. 
Specifically, we show that for an \ac{NS} decoding strategy with $100$ ($1,000$) samples, our method outperforms the training solutions on average for instance sizes of $200$ ($100$) customers and above. 
Figure \ref{fig:hgs_1_vrpt5_1000} further illustrates this point using an instance size of $400$ customers as an example. The figure compares the solutions from \ac{FM-MCVRP} under an \ac{NS} decoding strategy with $1,000$ samples to the solutions obtained from running \ac{HGS} once per problem instance with a 5 second time limit.
\ac{FM-MCVRP} solutions are shifted to the left compared to the \ac{HGS} solutions with a mean relative difference of $-1.05\%$. This improvement over the \ac{HGS} solutions is statistically significant as confirmed by a one-sided paired samples t-test \citep{Ross2017PairedT-Test} (see Appendix \ref{app:test-stats}).

Lastly, we note that for large problem instances, \ac{FM-MCVRP} yields competitive solutions compared to a state-of-the-art heuristic such as \ac{HGS}, when held to similarly restrictive computational time constraints. 
For instance, under a \ac{GS} decoding strategy, \ac{FM-MCVRP} finds solutions for 400-customer instances after around 6 seconds that are on average within 1.96\% of the solutions obtained by \ac{HGS} with a 5 second time limit.

\begin{table}[htbp]
    \tiny
    \begin{tabularx}{\textwidth}{l|X|cc|cc|c}
        \toprule 
        & & \multicolumn{2}{c|}{\bf Obj.} & \multicolumn{2}{c|}{\bf Gap (\%)} & {\bf Time} \\
        {\bf N} & {\bf Method (Decoder)} & Avg. & $80\%$ IP Range & Avg. & $80\%$ IP Range & Avg. \\
        \midrule
        \multirow{5}{*}{20} 
        & \ac{HGS} (no decoding, $s=1$) & 5.01 & 4.42 -- 5.62 & \multicolumn{2}{c|}{\it --- baseline --- } & 5.00s \\
         & \ac{FM-MCVRP} (\ac{GS}) & 5.42 & 4.73 -- 6.12 & 8.29 & 0.35 -- 17.25 & 0.26s \\
         & \ac{FM-MCVRP} (\ac{NS}, $s=1$) & 5.45 & 4.74 -- 6.19 & 9.05 & 0.32 -- 19.98 & 0.26s \\
         & \ac{FM-MCVRP} (\ac{NS}, $s=100$) & 5.09 & 4.48 -- 5.70 & 1.62 & 0.00 -- 5.01 & 2.48s \\
         & \ac{FM-MCVRP} (\ac{NS}, $s=1000$) & 5.07 & 4.48 -- 5.68 & 1.18 & 0.00 -- 4.11 & 24.77s \\
        \midrule
        \multirow{5}{*}{50}
         & \ac{HGS} (no decoding, $s=1$) & 8.35 & 7.69 -- 9.02 & \multicolumn{2}{c|}{\it --- baseline --- } & 5.00s \\
         & \ac{FM-MCVRP} (\ac{GS}) & 8.78 & 8.08 -- 9.55 & 5.21 & 0.89 -- 9.91 & 0.55s \\
         & \ac{FM-MCVRP} (\ac{NS}, $s=1$) & 8.79 & 8.07 -- 9.58 & 5.27 & 0.92 -- 10.50 & 0.55s \\
         & \ac{FM-MCVRP} (\ac{NS}, $s=100$) & 8.46 & 7.83 -- 9.13 & 1.30 & -0.68 -- 3.69 & 7.67s \\
         & \ac{FM-MCVRP} (\ac{NS}, $s=1000$) & 8.42 & 7.79 -- 9.09 & 0.87 & -0.90 -- 3.10 & 1.28min \\
        \midrule
        \multirow{5}{*}{100}
         & \ac{HGS} (no decoding, $s=1$) & 12.65 & 11.93 -- 13.43 & \multicolumn{2}{c|}{\it --- baseline --- } & 5.00s \\
         & \ac{FM-MCVRP} (\ac{GS}) & 13.13 & 12.31 -- 14.02 & 3.75 & 0.38 -- 7.36 & 1.03s \\
         & \ac{FM-MCVRP} (\ac{NS}, $s=1$) & 13.13 & 12.29 -- 14.01 & 3.75 & 0.49 -- 7.11 & 1.03s \\
         & \ac{FM-MCVRP} (\ac{NS}, $s=100$) & 12.68 & 11.96 -- 13.46 & 0.21 & -1.51 -- 2.10 & 24.43s \\
         & \ac{FM-MCVRP} (\ac{NS}, $s=1000$) & 12.61 & 11.90 -- 13.38 & -0.32 & -1.99 -- 1.25 & 4.07min \\
        \midrule
        \multirow{5}{*}{200}
         & \ac{HGS} (no decoding, $s=1$) & 19.77 & 18.76 -- 20.80 & \multicolumn{2}{c|}{\it --- baseline --- } & 5.00s \\
         & \ac{FM-MCVRP} (\ac{GS}) & 20.27 & 19.15 -- 21.38 & 2.50 & 0.06 -- 5.07 & 2.13s \\
         & \ac{FM-MCVRP} (\ac{NS}, $s=1$) & 20.27 & 19.16 -- 21.40 & 2.50 & 0.18 -- 4.84 & 2.13s \\
         & \ac{FM-MCVRP} (\ac{NS}, $s=100$) & 19.65 & 18.64 -- 20.69 & -0.61 & -1.89 -- 0.67 & 1.53min \\
         & \ac{FM-MCVRP} (\ac{NS}, $s=1000$) & 19.54 & 18.55 -- 20.55 & -1.16 & -2.38 -- 0.01 & 15.30min \\
        \midrule
        \multirow{5}{*}{400}
         & \ac{HGS} (no decoding, $s=1$) & 35.08 & 33.24 -- 36.81 & \multicolumn{2}{c|}{\it --- baseline --- } & 5.00s \\
         & \ac{FM-MCVRP} (\ac{GS}) & 35.76 & 33.82 -- 37.67 & 1.96 & 0.15 -- 4.09 & 5.97s \\
         & \ac{FM-MCVRP} (\ac{NS}, $s=1$) & 35.80 & 33.81 -- 37.77 & 2.05 & 0.18 -- 4.04 & 5.97s \\
         & \ac{FM-MCVRP} (\ac{NS}, $s=100$) & 34.89 & 33.08 -- 36.65 & -0.52 & -1.35 -- 0.32 & 6.31min \\
         & \ac{FM-MCVRP} (\ac{NS}, $s=1000$) & 34.71 & 32.87 -- 36.45 & -1.05 & -1.82 -- -0.29 & 63.14min \\
        \midrule
        \multicolumn{7}{l}{
            \tiny 
            N: instance size; 
            IP: inter-percentile;
            \ac{GS}: \acl{GS};
            \ac{NS}: \acl{NS};
            $s$: number of samples
            }  \\
        \multicolumn{7}{l}{
            \tiny 
            Metrics are reported over $m=1,000$ instances per instance size
            }  \\
        \multicolumn{7}{l}{
            \tiny 
            Times are reported based on average time per instance
            }  \\
        \bottomrule
    \end{tabularx}%
    \caption{Performance of \ac{FM-MCVRP} trained on 38.1M single-sample solutions from \ac{HGS} under a 5-second time limit compared to its training data for various decoding strategies.}
    \label{table:key_result}
\end{table}

\subsection{Solution Quality Under a Less Restrictive Computational Budget}
\label{sec:distance_distribution}

\begin{figure}[htbp]
    \centering
    \includegraphics[width=.75\textwidth]{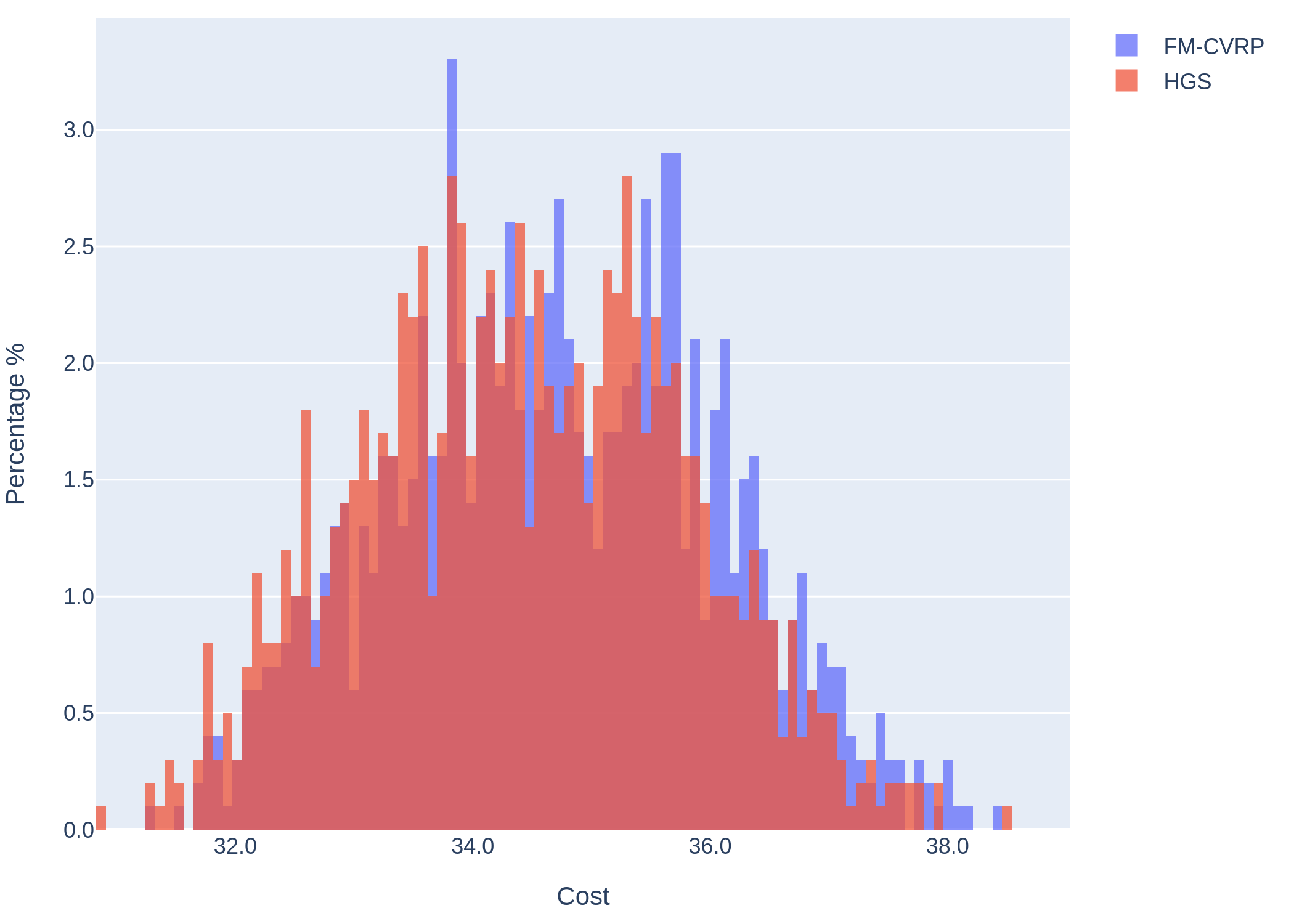}        
    \caption{Distance distributions of \ac{FM-MCVRP} (\ac{NS}, 1,000 samples) and \ac{HGS} (best of 1,000 runs) on 1,000 400-customer problem instances.}
    \label{fig:hgs_1000_vrpt5_1000}
\end{figure}

A third key result from our analyses is that \ac{FM-MCVRP} trained on sub-optimal solutions produces competitive solutions to large problem instances compared to state-of-the-art heuristics, even when the computational budget is less constrained.
Figure \ref{fig:hgs_1000_vrpt5_1000} illustrates this finding using an instance size of $400$ customers as an example. 
It compares \ac{FM-MCVRP} solutions under an \ac{NS} decoding strategy with $1,000$ samples to the best out of $1,000$ \ac{HGS} runs per problem instance with a 5 second time limit per run.
\ac{FM-MCVRP} solutions are shifted to the right compared to the \ac{HGS} solutions with a mean relative difference of $0.81\%$.
This deterioration relative to the \ac{HGS} solutions is also statistically significant (see Appendix \ref{app:test-stats}).
Recall however, that our model has been trained on a set of sub-optimal solutions obtained from a single solution run of \ac{HGS} with a time limit of 5 seconds. 
Therefore, it is not surprising that our model is outperformed by the best solution out of 1,000 \ac{HGS} runs. However, given that \ac{FM-MCVRP} was trained on lower-quality solutions, it is noteworthy that the solutions it finds remain highly competitive.

Finally, as an illustration, we show in Figure \ref{fig:small_best_gap} (see Appendix \ref{app:example-solutions}) a 400-node problem instance where \ac{FM-MCVRP} outperformed \ac{HGS} within the sample size.

\subsection{Generalizing to Larger Problems}
\label{sec:generalizing_to_larger_problems}

Another key finding from our analyses is that our model generalizes well to problem instance sizes that were not part of the training data and go beyond the instance sizes that the model has seen during training. To illustrate this finding, we extend our performance comparison between \ac{FM-MCVRP} and \ac{HGS} to problem instances with $600$ and $800$ customers.
Since these larger problem instances require more time to decode, we only base our performance statistics on the solutions obtained for $100$ instances per instance size.

Table \ref{table:large_probs_results} shows that under a \ac{NS} decoding strategy with $100$ or $1,000$ samples, \ac{FM-MCVRP} (trained on solutions for 20 to 400-node problems obtained from running \ac{HGS} once per problem instance for 5 seconds, as discussed in Section \ref{sec:data}) continues to outperform single-run solutions from \ac{HGS} under a 5-second time limit. It yields inferior solutions compared to the single-run \ac{HGS} benchmark for a decoding strategy with only a single sample, however.  
Nonetheless, the general capability of our model to generalize to larger problem instances is noteworthy as it illustrates the applicability of our method to real-world scenarios due to its robustness to variations in the problem instance characteristics.
Moreover, this also indicates that it is possible to train \ac{FM-MCVRP} on a larger dataset of smaller instances solved to near-optimality, which is computationally less costly to generate, and then generalize to larger problem instances that one might be interested in solving.

Finally, as an illustration, we show in Figure \ref{fig:large_best_gap} (see Appendix \ref{app:example-solutions}) an 800-node problem instance where \ac{FM-MCVRP} outperformed \ac{HGS} within the sample size.

\begin{table}[htbp]
    \tiny
    \begin{tabularx}{\textwidth}{l|X|c|cc|cc|c}
        \toprule 
        & & & \multicolumn{2}{c|}{\bf Obj.} & \multicolumn{2}{c|}{\bf Gap (\%)} & {\bf Time} \\
        {\bf N} & {\bf Method (Decoder)} & {\bf Wins} & Avg. & $80\%$ IP Range & Avg. & $80\%$ IP Range & Avg. \\
        \midrule
        \multirow{7}{*}{600}
         & \ac{HGS} (no decoding, $s=1$) & - & 44.49 & 42.70 -- 46.30 & \multicolumn{2}{c|}{\it --- baseline --- } & 5.00s \\
         & \ac{FM-MCVRP} (GS, $s=1$) & 7 & 45.28 & 43.33 -- 47.39 & 1.78 & 0.30 -- 4.45 & 11.73s \\
         & \ac{FM-MCVRP} (NS, $s=1$) & 6 & 45.26 & 43.49 -- 47.14 & 1.72 & 0.24 -- 3.45 & 11.73s \\
        \cmidrule[0.1pt](){2-8}
         & \ac{HGS} (no decoding, $s=100$) & - & 43.91 & 41.95 -- 45.68 & \multicolumn{2}{c|}{\it --- baseline --- } & 12.75s \\
         & \ac{FM-MCVRP} (NS, $s=100$) & 5 & 44.23 & 42.25 -- 46.06 & 0.72 & 0.30 -- 1.16 & 14.52min \\
        \cmidrule[0.1pt](){2-8}
         & \ac{HGS} (no decoding, $s=1000$) & - & 43.75 & 41.71 -- 45.61 & \multicolumn{2}{c|}{\it --- baseline --- } & 2.13min \\
         & \ac{FM-MCVRP} (NS, $s=1000$) & 3 & 44.04 & 42.11 -- 45.92 & 0.66 & 0.29 -- 0.98 & 145.18min \\
        \midrule
        \multirow{7}{*}{800}
         & \ac{HGS} (no decoding, $s=1$) & - & 57.04 & 54.90 -- 59.83 & \multicolumn{2}{c|}{\it --- baseline --- } & 5.00s \\
         & \ac{FM-MCVRP} (GS, $s=1$) & 3 & 58.35 & 55.56 -- 61.03 & 2.29 & 0.64 -- 4.73 & 20.87s \\
         & \ac{FM-MCVRP} (NS, $s=1$) & 3 & 58.33 & 55.93 -- 61.61 & 2.26 & 0.51 -- 4.38 & 20.87s \\
        \cmidrule[0.1pt](){2-8}
         & \ac{HGS} (no decoding, $s=100$) & - & 56.46 & 54.38 -- 59.40 & \multicolumn{2}{c|}{\it --- baseline --- } & 12.83s \\
         & \ac{FM-MCVRP} (NS, $s=100$) & 3 & 56.91 & 54.56 -- 59.83 & 0.80 & 0.31 -- 1.32 & 27.35min \\
        \cmidrule[0.1pt](){2-8}
         & \ac{HGS} (no decoding, $s=1000$) & - & 56.29 & 54.12 -- 59.28 & \multicolumn{2}{c|}{\it --- baseline --- } & 2.14min \\
         & \ac{FM-MCVRP} (NS, $s=1000$) & 1 & 56.69 & 54.34 -- 59.81 & 0.71 & 0.36 -- 1.06 & 273.53min \\
        \midrule
        \multicolumn{8}{l}{
            \tiny 
            N: instance size; 
            IP: inter-percentile;
            \ac{GS}: \acl{GS};
            \ac{NS}: \acl{NS};
            $s$: number of samples
            }  \\
        \multicolumn{8}{l}{
            \tiny 
            Metrics are reported over $m=1,000$ instances per instance size
            }  \\
        \multicolumn{8}{l}{
            \tiny 
            Wins represent number problem instances in which the given method outperforms the benchmark method
            }  \\
        \multicolumn{8}{l}{
            \tiny 
            Times are reported based on average time per instance
            }  \\
        \bottomrule
    \end{tabularx}%
    \caption{Comparison of \ac{FM-MCVRP} with \ac{HGS} as the baseline for problem instances with 600 and 800 customers. The gap \% is computed with respect to \ac{HGS} within the same sample size.}
    \label{table:large_probs_results}
\end{table}

\subsection{Comparison with Other State-of-the-Art Methods}
\label{sec:extended_comparison}

In this section, we extend our performance analysis and compare \ac{FM-MCVRP} to two other state-of-the-art methods that solve the \ac{CVRP}.
First, we present a comparison with \ac{LKH-3}, which is the best-performing heuristic for instance sizes of $400$ customers or less across our experiments.
Second, we present a comparison with \ac{AM}, which is a widely discussed \ac{DRL} approach to solving the \ac{CVRP}.

\subsubsection{Comparison with \ac{LKH-3}.}
\label{sec:LKH3_comp}

After demonstrating that \ac{FM-MCVRP} generalizes to better solution qualities, larger problem instances, and less resource constrained heuristic solutions in the previous sections, we now want to show how \ac{FM-MCVRP} solutions compare to the solutions found by the best-in-class heuristic, which is arguably \ac{LKH-3} for problem instances with up to $400$ customers.
The corresponding results from our numerical experiments are summarized in Table \ref{table:fmcvrp_vs_lkh3_baseline}, which groups the obtained solutions from \ac{LKH-3} (baseline), \ac{HGS}, and \ac{FM-MCVRP} by instance size and the number of samples / algorithm runs.

As Table \ref{table:fmcvrp_vs_lkh3_baseline} shows, the single-sample solutions from \ac{FM-MCVRP} (for both \ac{GS} and \ac{NS}) are inferior to the single-run solutions obtained from \ac{LKH-3} across all instance sizes, with an average relative difference in the solution value of up to $8.59\%$ for 20-customer problems. However, there are two effects worth mentioning in our numerical results, which are affecting the relative performance of our method compared to the benchmark simultaneously.
First, similar to what we have seen in Section \ref{sec:outperform_heuristic} in comparison to \ac{HGS}, the relative performance gap between \ac{LKH-3} (single-run) and \ac{FM-MCVRP} (single-sample) reduces rapidly as problem instances get larger, which indicates that the solution quality of \ac{LKH-3} deteriorates more rapidly than that of our method as instances get larger.
This is most clearly visible in Table \ref{table:fmcvrp_vs_lkh3_baseline} when comparing the single-run solutions from \ac{LKH-3} with their single-sample counterparts from \ac{FM-MCVRP}. Here the average relative difference in solution value falls from $8.59\%$ for 20-customer problems to $3.02\%$ for 400-customer problems.
Second, as we increase the number of samples / runs, \ac{FM-MCVRP} solutions quickly become more competitive. However, the magnitude of this effect is dampened by the first effect. For 20-customer problem instances, the average relative difference in the solution value between \ac{LKH-3} and \ac{FM-MCVRP} drops to $1.65\%$ and $1.24\%$ for $100$ and $1,000$ runs / samples, respectively. This corresponds to a gap reduction by over $80\%$ and over $85\%$, respectively. For 400-customer problem, the corresponding gap reductions from increasing the number of runs / samples from one to $100$ and $1,000$ is only around $28\%$ and $33\%$, respectively.

\begin{table}[htbp]
    \tiny
    \begin{tabularx}{\textwidth}{l|X|c|cc|cc|c}
        \toprule 
        & & & \multicolumn{2}{c|}{\bf Obj.} & \multicolumn{2}{c|}{\bf Gap (\%)} & {\bf Time} \\
        {\bf N} & {\bf Method (Decoder)} & {\bf Wins} & Avg. & $80\%$ IP Range & Avg. & $80\%$ IP Range & Avg. \\
        \midrule
        \multirow{10}{*}{20}
         & \ac{LKH-3} (no decoding, $s=1$) & - & 5.01 & 4.43 -- 5.63 & \multicolumn{2}{c|}{\it --- baseline --- } & 0.02s \\
         & \ac{HGS} (no decoding, $s=1$) & 359 & 5.01 & 4.42 -- 5.62 & -0.04 & -0.26 -- 0.05 & 5.00s \\
         & \ac{FM-MCVRP} (GS, $s=1$) & 38 & 5.42 & 4.73 -- 6.12 & 8.18 & 0.23 -- 17.21 & 0.26s \\
         & \ac{FM-MCVRP} (NS, $s=1$) & 33 & 5.44 & 4.74 -- 6.16 & 8.59 & 0.26 -- 19.30 & 0.26s \\
        \cmidrule[0.1pt](){2-8}
         & \ac{LKH-3} (no decoding, $s=100$) & - & 5.01 & 4.43 -- 5.61 & \multicolumn{2}{c|}{\it --- baseline --- } & 0.20s \\
         & \ac{HGS} (no decoding, $s=100$) & 73 & 5.00 & 4.42 -- 5.61 & -0.08 & 0.00 -- 0.00 & 11.16s \\
         & \ac{FM-MCVRP} (NS, $s=100$) & 36 & 5.09 & 4.49 -- 5.69 & 1.65 & 0.00 -- 5.06 & 2.48s \\
        \cmidrule[0.1pt](){2-8}
         & \ac{LKH-3} (no decoding, $s=1000$) & - & 5.01 & 4.43 -- 5.61 & \multicolumn{2}{c|}{\it --- baseline --- } & 2.03s \\
         & \ac{HGS} (no decoding, $s=1000$) & 72 & 5.00 & 4.42 -- 5.61 & -0.08 & 0.00 -- 0.00 & 1.85min \\
         & \ac{FM-MCVRP} (NS, $s=1000$) & 45 & 5.07 & 4.48 -- 5.68 & 1.24 & 0.00 -- 4.19 & 24.77s \\
        \midrule
        \multirow{10}{*}{50}
         & \ac{LKH-3} (no decoding, $s=1$) & - & 8.33 & 7.70 -- 9.02 & \multicolumn{2}{c|}{\it --- baseline --- } & 0.08s \\
         & \ac{HGS} (no decoding, $s=1$) & 369 & 8.35 & 7.72 -- 8.99 & 0.18 & -1.45 -- 1.89 & 5.00s \\
         & \ac{FM-MCVRP} (GS, $s=1$) & 36 & 8.78 & 8.08 -- 9.55 & 5.42 & 1.03 -- 10.25 & 0.55s \\
         & \ac{FM-MCVRP} (NS, $s=1$) & 32 & 8.80 & 8.08 -- 9.56 & 5.61 & 1.14 -- 10.83 & 0.55s \\
        \cmidrule[0.1pt](){2-8}
         & \ac{LKH-3} (no decoding, $s=100$) & - & 8.26 & 7.65 -- 8.89 & \multicolumn{2}{c|}{\it --- baseline --- } & 0.43s \\
         & \ac{HGS} (no decoding, $s=100$) & 117 & 8.26 & 7.64 -- 8.89 & -0.03 & 0.00 -- 0.00 & 11.52s \\
         & \ac{FM-MCVRP} (NS, $s=100$) & 9 & 8.46 & 7.82 -- 9.13 & 2.40 & 0.37 -- 4.76 & 7.67s \\
        \cmidrule[0.1pt](){2-8}
         & \ac{LKH-3} (no decoding, $s=1000$) & - & 8.26 & 7.65 -- 8.89 & \multicolumn{2}{c|}{\it --- baseline --- } & 4.28s \\
         & \ac{HGS} (no decoding, $s=1000$) & 73 & 8.25 & 7.64 -- 8.89 & -0.04 & 0.00 -- 0.00 & 1.92min \\
         & \ac{FM-MCVRP} (NS, $s=1000$) & 11 & 8.42 & 7.79 -- 9.09 & 1.96 & 0.18 -- 4.11 & 1.28min \\
        \midrule
        \multirow{10}{*}{100}
         & \ac{LKH-3} (no decoding, $s=1$) & - & 12.53 & 11.78 -- 13.32 & \multicolumn{2}{c|}{\it --- baseline --- } & 0.47s \\
         & \ac{HGS} (no decoding, $s=1$) & 229 & 12.66 & 11.93 -- 13.43 & 1.02 & -1.00 -- 3.05 & 5.00s \\
         & \ac{FM-MCVRP} (GS, $s=1$) & 34 & 13.13 & 12.31 -- 14.02 & 4.77 & 1.30 -- 8.52 & 1.03s \\
         & \ac{FM-MCVRP} (NS, $s=1$) & 41 & 13.14 & 12.28 -- 14.01 & 4.84 & 1.34 -- 8.60 & 1.03s \\
        \cmidrule[0.1pt](){2-8}
         & \ac{LKH-3} (no decoding, $s=100$) & - & 12.32 & 11.63 -- 13.08 & \multicolumn{2}{c|}{\it --- baseline --- } & 0.99s \\
         & \ac{HGS} (no decoding, $s=100$) & 88 & 12.38 & 11.67 -- 13.15 & 0.47 & 0.00 -- 1.05 & 11.73s \\
         & \ac{FM-MCVRP} (NS, $s=100$) & 2 & 12.68 & 11.96 -- 13.44 & 2.87 & 1.20 -- 4.62 & 24.43s \\
        \cmidrule[0.1pt](){2-8}
         & \ac{LKH-3} (no decoding, $s=1000$) & - & 12.30 & 11.62 -- 13.07 & \multicolumn{2}{c|}{\it --- baseline --- } & 9.87s \\
         & \ac{HGS} (no decoding, $s=1000$) & 78 & 12.33 & 11.64 -- 13.08 & 0.23 & 0.00 -- 0.62 & 1.96min \\
         & \ac{FM-MCVRP} (NS, $s=1000$) & 4 & 12.61 & 11.90 -- 13.38 & 2.50 & 1.06 -- 4.09 & 4.07min \\
        \midrule
        \multirow{10}{*}{200}
         & \ac{LKH-3} (no decoding, $s=1$) & - & 19.48 & 18.44 -- 20.52 & \multicolumn{2}{c|}{\it --- baseline --- } & 0.91s \\
         & \ac{HGS} (no decoding, $s=1$) & 126 & 19.77 & 18.73 -- 20.81 & 1.48 & -0.29 -- 3.20 & 5.00s \\
         & \ac{FM-MCVRP} (GS, $s=1$) & 35 & 20.27 & 19.15 -- 21.38 & 4.05 & 1.40 -- 6.87 & 2.13s \\
         & \ac{FM-MCVRP} (NS, $s=1$) & 25 & 20.28 & 19.19 -- 21.43 & 4.12 & 1.34 -- 6.86 & 2.13s \\
        \cmidrule[0.1pt](){2-8}
         & \ac{LKH-3} (no decoding, $s=100$) & - & 19.13 & 18.13 -- 20.14 & \multicolumn{2}{c|}{\it --- baseline --- } & 2.75s \\
         & \ac{HGS} (no decoding, $s=100$) & 33 & 19.37 & 18.37 -- 20.39 & 1.25 & 0.54 -- 1.94 & 12.13s \\
         & \ac{FM-MCVRP} (NS, $s=100$) & 4 & 19.65 & 18.65 -- 20.69 & 2.76 & 1.62 -- 3.90 & 1.53min \\
        \cmidrule[0.1pt](){2-8}
         & \ac{LKH-3} (no decoding, $s=1000$) & - & 19.06 & 18.07 -- 20.07 & \multicolumn{2}{c|}{\it --- baseline --- } & 27.46s \\
         & \ac{HGS} (no decoding, $s=1000$) & 32 & 19.26 & 18.29 -- 20.26 & 1.06 & 0.52 -- 1.60 & 2.02min \\
         & \ac{FM-MCVRP} (NS, $s=1000$) & 3 & 19.54 & 18.55 -- 20.55 & 2.56 & 1.54 -- 3.52 & 15.30min \\
        \midrule
        \multirow{10}{*}{400}
         & \ac{LKH-3} (no decoding, $s=1$) & - & 34.70 & 32.89 -- 36.52 & \multicolumn{2}{c|}{\it --- baseline --- } & 3.88s \\
         & \ac{HGS} (no decoding, $s=1$) & 173 & 35.08 & 33.27 -- 36.79 & 1.12 & -0.64 -- 2.52 & 5.00s \\
         & \ac{FM-MCVRP} (GS, $s=1$) & 62 & 35.76 & 33.82 -- 37.67 & 3.08 & 0.73 -- 5.46 & 5.97s \\
         & \ac{FM-MCVRP} (NS, $s=1$) & 50 & 35.74 & 33.78 -- 37.61 & 3.02 & 0.70 -- 5.23 & 5.97s \\
        \cmidrule[0.1pt](){2-8}
         & \ac{LKH-3} (no decoding, $s=100$) & - & 34.16 & 32.37 -- 35.94 & \multicolumn{2}{c|}{\it --- baseline --- } & 10.93s \\
         & \ac{HGS} (no decoding, $s=100$) & 86 & 34.58 & 32.76 -- 36.35 & 1.25 & 0.15 -- 2.05 & 12.75s \\
         & \ac{FM-MCVRP} (NS, $s=100$) & 24 & 34.89 & 33.04 -- 36.68 & 2.16 & 0.84 -- 3.10 & 6.31min \\
        \cmidrule[0.1pt](){2-8}
         & \ac{LKH-3} (no decoding, $s=1000$) & - & 34.02 & 32.22 -- 35.80 & \multicolumn{2}{c|}{\it --- baseline --- } & 1.82min \\
         & \ac{HGS} (no decoding, $s=1000$) & 76 & 34.43 & 32.61 -- 36.19 & 1.21 & 0.24 -- 1.85 & 2.12min \\
         & \ac{FM-MCVRP} (NS, $s=1000$) & 31 & 34.71 & 32.87 -- 36.45 & 2.02 & 0.88 -- 2.81 & 63.14min \\
        \midrule
        \multicolumn{8}{l}{
            \tiny 
            N: instance size; 
            IP: inter-percentile;
            \ac{GS}: \acl{GS};
            \ac{NS}: \acl{NS};
            $s$: number of samples
            }  \\
        \multicolumn{8}{l}{
            \tiny 
            Metrics are reported over $m=1,000$ instances per instance size
            }  \\
        \multicolumn{8}{l}{
            \tiny 
            Wins represent number problem instances in which the given method outperforms the benchmark method
            }  \\
        \multicolumn{8}{l}{
            \tiny 
            Times are reported based on average time per instance
            }  \\
        \bottomrule
    \end{tabularx}%
    \caption{Comparison of \ac{FM-MCVRP} with \ac{LKH-3} as the baseline. The gap \% is computed with respect to \ac{LKH-3} within the same sample size.}
    \label{table:fmcvrp_vs_lkh3_baseline}
\end{table}

\subsubsection{Comparison with \ac{AM}.}
\label{sec:AM_comp}

After comparing \ac{FM-MCVRP} with state-of-the-art heuristics, we also want to assess its performance relative to a recent, and widely discussed \ac{DRL} approach to routing, \ac{AM}.
Table \ref{table:fmcvrp_vs_am_baseline} shows the corresponding results of our numerical analyses grouped by instance size and the number of samples considered during decoding. Here, we make a number of important observations.

First, and most notably, \ac{AM} diverges and thus fails to produce meaningful solutions for problem instances with 400 and more customers.

Second, \ac{FM-MCVRP} solutions frequently and consistently outperform the solutions obtained from \ac{AM}. For sample sizes of $100$ and $1,000$, the solution values found by \ac{FM-MCVRP} are up to $2.07\%$ better on average than those found by \ac{AM}. For large problem instances (here, 200 customers) the $80\%$ inter-percentile range of the relative gap of \ac{FM-MCVRP} solution over the \ac{AM} solution spans from $-3.23\%$ to $-0.88\%$, indicating that our method almost always outperforms \ac{AM}.
Only for relatively small problem instances (50 customers or less) and a sample size of one, \ac{FM-MCVRP} on average yields worse solutions than \ac{AM}. However, even in these cases, the $80\%$ inter-percentile range of the relative gap spans well into the negative range, indicating that \ac{FM-MCVRP} still regularly beats \ac{AM}.

It is important to note that these results were obtained for \ac{FM-MCVRP} being trained on sub-optimal solutions obtained from \ac{HGS} under a strict time limit (see Section \ref{sec:experiment_setup}), while \ac{AM} was retrained for this comparison under a much more generous training regime (see Appendix \ref{app:retraining-am}).
Further, we note that the \ac{AM} is not a unified model. Unlike \ac{FM-MCVRP}, which as we show above generalizes well to unseen instance sizes and vehicle capacities, we need to train \ac{AM} for every specific combination of instance size and vehicle capacity we want to apply it to.

\begin{table}[htbp]
    \tiny
    \begin{tabularx}{\textwidth}{l|X|c|cc|cc|c}
        \toprule 
        & & & \multicolumn{2}{c|}{\bf Obj.} & \multicolumn{2}{c|}{\bf Gap (\%)} & {\bf Time} \\
        {\bf N} & {\bf Method (Decoder)} & {\bf Wins} & Avg. & $80\%$ IP Range & Avg. & $80\%$ IP Range & Avg. \\
        \midrule
        \multirow{8}{*}{20}
         & \ac{AM} (VS, $s=1$) & - & 5.17 & 4.54 -- 5.84 & \multicolumn{2}{c|}{\it --- baseline --- } & 0.03s \\
         & \ac{FM-MCVRP} (GS, $s=1$) & 297 & 5.42 & 4.73 -- 6.12 & 5.06 & -2.69 -- 14.84 & 0.26s \\
         & \ac{FM-MCVRP} (NS, $s=1$) & 303 & 5.44 & 4.74 -- 6.16 & 5.46 & -2.61 -- 15.80 & 0.26s \\
        \cmidrule[0.1pt](){2-8}
         & \ac{AM} (VS, $s=100$) & - & 5.09 & 4.47 -- 5.73 & \multicolumn{2}{c|}{\it --- baseline --- } & 0.03s \\
         & \ac{FM-MCVRP} (NS, $s=100$) & 539 & 5.09 & 4.49 -- 5.69 & -0.01 & -2.85 -- 3.31 & 2.48s \\
        \cmidrule[0.1pt](){2-8}
         & \ac{AM} (VS, $s=1000$) & - & 5.08 & 4.46 -- 5.72 & \multicolumn{2}{c|}{\it --- baseline --- } & 0.03s \\
         & \ac{FM-MCVRP} (NS, $s=1000$) & 552 & 5.07 & 4.48 -- 5.68 & -0.15 & -2.52 -- 2.48 & 24.77s \\
        \midrule
        \multirow{8}{*}{50}
         & \ac{AM} (VS, $s=1$) & - & 8.74 & 8.04 -- 9.47 & \multicolumn{2}{c|}{\it --- baseline --- } & 0.06s \\
         & \ac{FM-MCVRP} (GS, $s=1$) & 510 & 8.78 & 8.08 -- 9.55 & 0.54 & -4.27 -- 5.98 & 0.55s \\
         & \ac{FM-MCVRP} (NS, $s=1$) & 473 & 8.80 & 8.08 -- 9.56 & 0.72 & -4.11 -- 6.39 & 0.55s \\
        \cmidrule[0.1pt](){2-8}
         & \ac{AM} (VS, $s=100$) & - & 8.53 & 7.89 -- 9.19 & \multicolumn{2}{c|}{\it --- baseline --- } & 0.07s \\
         & \ac{FM-MCVRP} (NS, $s=100$) & 674 & 8.46 & 7.82 -- 9.13 & -0.78 & -3.36 -- 1.93 & 7.67s \\
        \cmidrule[0.1pt](){2-8}
         & \ac{AM} (VS, $s=1000$) & - & 8.49 & 7.85 -- 9.16 & \multicolumn{2}{c|}{\it --- baseline --- } & 0.08s \\
         & \ac{FM-MCVRP} (NS, $s=1000$) & 694 & 8.42 & 7.79 -- 9.09 & -0.77 & -3.09 -- 1.44 & 1.28min \\
        \midrule
        \multirow{8}{*}{100}
         & \ac{AM} (VS, $s=1$) & - & 13.15 & 12.34 -- 13.99 & \multicolumn{2}{c|}{\it --- baseline --- } & 0.12s \\
         & \ac{FM-MCVRP} (GS, $s=1$) & 540 & 13.13 & 12.31 -- 14.02 & -0.15 & -4.07 -- 3.66 & 1.03s \\
         & \ac{FM-MCVRP} (NS, $s=1$) & 533 & 13.14 & 12.28 -- 14.01 & -0.07 & -3.90 -- 3.74 & 1.03s \\
        \cmidrule[0.1pt](){2-8}
         & \ac{AM} (VS, $s=100$) & - & 12.82 & 12.08 -- 13.64 & \multicolumn{2}{c|}{\it --- baseline --- } & 0.13s \\
         & \ac{FM-MCVRP} (NS, $s=100$) & 751 & 12.68 & 11.96 -- 13.44 & -1.08 & -3.08 -- 1.08 & 24.43s \\
        \cmidrule[0.1pt](){2-8}
         & \ac{AM} (VS, $s=1000$) & - & 12.75 & 12.02 -- 13.57 & \multicolumn{2}{c|}{\it --- baseline --- } & 0.17s \\
         & \ac{FM-MCVRP} (NS, $s=1000$) & 758 & 12.61 & 11.90 -- 13.38 & -1.06 & -2.91 -- 0.85 & 4.07min \\
        \midrule
        \multirow{8}{*}{200}
         & \ac{AM} (VS, $s=1$) & - & 20.59 & 19.55 -- 21.62 & \multicolumn{2}{c|}{\it --- baseline --- } & 0.29s \\
         & \ac{FM-MCVRP} (GS, $s=1$) & 782 & 20.27 & 19.15 -- 21.38 & -1.54 & -4.17 -- 1.21 & 2.13s \\
         & \ac{FM-MCVRP} (NS, $s=1$) & 781 & 20.28 & 19.19 -- 21.43 & -1.46 & -4.10 -- 1.25 & 2.13s \\
        \cmidrule[0.1pt](){2-8}
         & \ac{AM} (VS, $s=100$) & - & 20.07 & 19.08 -- 21.08 & \multicolumn{2}{c|}{\it --- baseline --- } & 0.31s \\
         & \ac{FM-MCVRP} (NS, $s=100$) & 979 & 19.65 & 18.65 -- 20.69 & -2.07 & -3.40 -- -0.75 & 1.53min \\
        \cmidrule[0.1pt](){2-8}
         & \ac{AM} (VS, $s=1000$) & - & 19.95 & 18.97 -- 20.95 & \multicolumn{2}{c|}{\it --- baseline --- } & 0.53s \\
         & \ac{FM-MCVRP} (NS, $s=1000$) & 981 & 19.54 & 18.55 -- 20.55 & -2.02 & -3.23 -- -0.88 & 15.30min \\
        \midrule
        \multirow{8}{*}{400}
         & \ac{AM} (VS, $s=1$) & - & - & - & - & - \\
         & \ac{FM-MCVRP} (GS, $s=1$) & 1000 & 35.76 & 33.82 -- 37.67 & - & - & 5.97s \\
         & \ac{FM-MCVRP} (NS, $s=1$) & 1000 & 35.74 & 33.78 -- 37.61 & - & - & 5.97s \\
        \cmidrule[0.1pt](){2-8}
         & \ac{AM} (VS, $s=100$) & - & - & - & - & - \\
         & \ac{FM-MCVRP} (NS, $s=100$) & 1000 & 34.89 & 33.04 -- 36.68 & - & - & 6.31min \\
        \cmidrule[0.1pt](){2-8}
         & \ac{AM} (VS, $s=1000$) & - & - & - & - & - \\
         & \ac{FM-MCVRP} (NS, $s=1000$) & 1000 & 34.71 & 32.87 -- 36.45 & - & - & 63.14min \\
        \midrule
        \multicolumn{8}{l}{
            \tiny 
            N: instance size; 
            IP: inter-percentile;
            VS: vanilla sampling;
            \ac{GS}: \acl{GS};
            \ac{NS}: \acl{NS};
            $s$: number of samples
            }  \\
        \multicolumn{8}{l}{
            \tiny 
            Metrics are reported over $m=1,000$ instances per instance size
            }  \\
        \multicolumn{8}{l}{
            \tiny 
            Times are reported based on average time per instance
            }  \\
        \bottomrule
    \end{tabularx}%
    \caption{Comparison of \ac{FM-MCVRP} with \ac{AM} as the baseline. The gap \% is computed with respect to \ac{AM} within the same sample size.}
    \label{table:fmcvrp_vs_am_baseline}
\end{table}

\section{Managerial Implications}
\label{sec:managerial_insights}
In the increasingly complex world of delivery logistics, optimizing the \ac{CVRP} with state-of-the-art methods can yield substantial dividends.
Specifically, consider how a 1\% improvement in distance travelled for a global logistics company can have significant cost savings.
In this paper, we leverage state-of-the-art advancements in \acp{LLM} and offer potentially transformative solutions for real-world logistics challenges.
We outline four insights next.

\paragraph{Supervised Learning on Historical Data.}
Most, if not all global logistics companies have dedicated \ac{OR} teams and use sophisticated algorithms that have solved the \ac{CVRP} over the past few decades.
These companies likely have a vast amount of historical problem-solution pairs.
In addition, many of these companies may also have records of how routes were executed by the driver in reality, likely taking into account other factors beyond total travel distance, time, or cost. These factors could include safety, convenience, and other factors.
We show that our supervised learning method is effective in learning from a state-of-the-art heuristic solutions to the \ac{CVRP}, and we hypothesize that it can also effectively learn from real-world generated solutions that incorporate more complex objectives and constraints followed and adhered to by actual drivers.
With large amounts of historical data readily available, these companies can implement our method to improve their route operations and potentially learn from the tacit knowledge of their most experienced and productive drivers.

\paragraph{Unified Model for Varying Numbers of Customers and Truck Capacities.}
Our model also provides significantly more convenience from a \ac{MLOps} perspective.
As mentioned above, prior work based on \ac{DRL} require a specific model to be trained for a given number of customers and truck capacity.
However, this is not a given in real-world delivery problems.
Using these methods would require \ac{MLOps} teams to deploy and maintain multiple models tailored for specific customer counts and truck capacities.
Our approach introduces a unified model that performs well over a wide spectrum of problem sizes and vehicle capacities, simplifying the deployment process and reducing model maintenance effort.

\paragraph{Generalizing to Superior Solution Qualities.}
Beyond mere adaptability to different customer sizes and truck capacities, our model is able to decode solutions of higher quality than what it was trained on.
This property could allow our proposed model to continuously improve the route quality of a company, as it learns from historical or algorithmically generated solutions, and subsequently proposes higher quality solutions that could themselves be used to further (re-)train our model.

\paragraph{Scalability Beyond Normal Operations.}
Another significant finding of our research is our model's capability to handle larger problem sizes than those it was initially trained on.
In particular, we trained the model on 20 to 400-node problem instances, and found that the model could still produce solutions of high quality for 600 and 800-node problem instances. 
In practical terms, a company can train a model on data for regular delivery scenarios and can confidently apply this model to peak demand periods, even if the model has not previously encountered such high volumes during training.

All in all, our findings chart a path for delivery companies to embrace \ac{ML}-based routing methods.
Especially as the e-commerce landscape rapidly evolves, harnessing \ac{ML} methods for route planning could be critical for companies to make their delivery operations more flexible, customer-centric, scalable, and adaptable -- essential prerequisites for driving down costs and enhancing overall customer satisfaction.


\section{Conclusion}
\label{sec:conclusion}

In this paper, we propose the \ac{FM-MCVRP}, a novel \acf{DL} model that solves the so-called \acf{MCVRP}, a variant of the \ac{CVRP} that closely mimics real-world delivery problems.
To the best of our knowledge, our work is the first to leverage Transformers in an \ac{LLM} framework to solve the \ac{MCVRP}, contrary to recent works that use the \acf{PN} framework.

Our proposed unified model and the findings from our numerical study, which demonstrate competitiveness with state-of-the-art heuristics, are of high significance to the academic community as they constitute a first step towards successfully applying \ac{LLM} frameworks to \ac{CO} problems. 
They are also of high significance to industrial practice as many real-world delivery problems operate within fixed and given operational environments (i.e., known road networks, customer addresses).
The type of model presented in this paper can help to exploit patterns in these environments, learn from existing operational data, and gradually improve over previously found solutions.

The main limitations of our proposed work are threefold.
First, we intentionally trained \ac{FM-MCVRP} with sub-optimal solutions as we wanted to show that \ac{FM-MCVRP} is able to generate solutions of a higher quality than the sub-optimal solutions it was trained on.
Future research should explore using higher quality solutions by potentially extending the 5s time limit, and exploring the limit in which outperformance is no longer possible.

Second, \ac{FM-MCVRP} was trained with the \ac{T5} schedule, which consists of a fixed learning rate for a certain number of steps followed by an exponential decay of the learning rate.
However, \citet{Iyer2023Wide-minimaSchedule} recently proposed the \emph{Knee} training schedule and showed that this schedule increases the performance of the model.
Future work should leverage this work and similar insights into model training to potentially obtain better models.

Third, as \ac{FM-MCVRP} is a conditional probability model, autoregressive decoding is required to obtain solutions and this operation cannot be parallelized.
Future work could involve a non-autoregressive model that can be parallelized, thus speeding up the decoding process significantly.

There are a number of additional areas for future research that appear particularly promising.
First, as seen from the 2021 Amazon Last Mile Routing Challenge \citep{Merchan20222021Set}, favorable solutions to real-world routing problems are often not distance optimal. 
Instead, drivers optimize for more complex objective functions, aiming at balancing safety, convenience, and other factors beyond cost efficiency and our proposed supervised learning method is can potentially capture these preferences.

Second, while \citet{Raffel2020ExploringTransformer} conclude that the encoder-decoder architecture works best for the type of problem we are solving in this paper, it would be worth exploring \ac{GPT}-style (decoder-only) architectures. While this would result in significantly higher computational cost, a \ac{GPT}-style architecture would have access to the features of all layers in the Transformer, which potentially enables learning embeddings with better representations.

Lastly, a particularly intriguing area of future research is the development of a system in which a \ac{DL} model starts by learning from a state-of-the-art heuristic and then produces increasingly better solutions that can be bootstrapped into training the model itself, leading to even better solutions, and creating a positive feedback loop.


%
%
%
\newpage
\begin{APPENDICES}

\section{Attention Mechanism}
\label{app:attention}
At a high level, the Attention mechanism proposed by \cite{Vaswani2017AttentionNeed} takes a set of feature vectors in the form of a matrix and transforms it into another matrix of higher-level features of the same size. Specifically, the equations are given by
\begin{align}
    \text{Attention}(Q, K, V) = \text{softmax}\left(\frac{QK^T}{\sqrt{d_k}}\right) V, \label{eq:attention}
\end{align}
with $\text{softmax}(x_{i}) = \frac{\exp(x_i)}{\sum_j \exp(x_j)}$, and $Q = XW_Q$,  $K = XW_K$, and $V = XW_V$,
where $X$ is the input matrix containing the feature vectors; 
$W_Q$, $W_K$, and $W_V$ are learned weight matrices responsible for transforming the input $X$ into Query (Q), Key (K), and Value (V) matrices, respectively;
$QK^T$ is the dot product between the query and key matrices, resulting in a matrix whose element at the $i$-th row and $j$-th column represents the compatibility of the $i$-th query with the $j$-th key;
$\sqrt{d_k}$ is a scaling factor, where $d_k$ is the dimensionality of the queries and keys. This scaling ensures that the magnitudes of the dot products do not grow too large, which could potentially lead to gradients that are difficult to manage during the optimization process.
In words, this mechanism allows the model to decide which aspects of the input to focus on (via $QK^T$) and which parts should be summed in the final output (via $V$).

\section{Choosing Hyperparameters}
\label{app:choosing_hyperparams}
Modern \ac{DL} approaches, including Transformer models, are predominantly trained with \ac{SGD} \citep{Robbins1951AMethod}.
In \ac{SGD}, there are three high-level decisions that have to be made: the type of optimizer to use, setting the learning rate, and setting the batch size.

\paragraph{Optimizers.}
In \ac{SGD}, the most basic form of the parameter update rule is $\theta_{t+1} = \theta_t - \alpha g_t$, where $\theta_t$ and $g_t$ represent the parameters and the gradient respectively at time step $t$, and $\alpha$ the learning rate in this specific step.
There are a myriad of optimizers in the \ac{DL} literature and most if not all of them adapt this basic parameter update rule.
We focus our discussion on three optimizers that are commonly used in \ac{DL} research: 
First, the \emph{Adam} optimizer \citep{Kingma2014Adam:Optimization} is a popular optimizer that was commonly used in early \ac{DL} research due to its fast convergence when tested empirically.
Second, \emph{AdamW} \citep{Loshchilov2019DecoupledRegularization} improves on Adam with a parameter update rule that decouples the weight decay and produces models that generalize better when compared with Adam.
Lastly, \emph{AdaFactor} \citep{Shazeer2018Adafactor:Cost} is largely equivalent to Adam but is more memory efficient and achieves results comparable to models trained with Adam.

All three of these optimizers generally involve the first and second moments of the gradients.
The corresponding parameters, $\beta_1$ and $\beta_2$, can be tuned along with $\alpha$. 
In practice, the default implementation of AdamW in PyTorch sets the values of $\alpha$, $\beta_1$ and $\beta_2$ to $0.001$, $0.900$ and $0.999$ respectively.

\paragraph{Learning rates.}
Depending on the type of \ac{DL} model, the initial learning rate is another hyperparameter that can be tuned.
We omit the discussion of setting the initial learning rate as it is generally a number set to the default of $10^{-3}$ in state-of-the-art \ac{ML} libraries like PyTorch \citep{Paszke2019Pytorch:Library} and TensorFlow \citep{MartinAbadi2015Systems}, or determined based on a grid search.
We instead focus our discussion on two broad topics: the warm-up schedule and the annealing schedule.
\cite{Goyal2017AccurateHour} first proposed a gradual warm-up, where the learning rate starts from zero and gradually increases to the desired learning rate.
\citet{Goyal2017AccurateHour} argue that this enables healthy convergence at the start of training.
Their warm-up schedule is commonly used by most recently proposed \ac{DL} models.
For instance, \ac{LLaMA-2}, a popular \ac{LLM}, uses the gradual warm-up scheme for the first 2,000 steps of training \citep{Touvron2023LlamaModels}.
In contrast, the \ac{T5} model, another relatively popular \ac{LLM}, does not follow a gradual warm-up, but instead starts with a high constant learning rate \citep{Raffel2020ExploringTransformer}.
Regardless of the warm-up schedule, after reaching the peak learning rate, most models follow an annealing schedule that gradually reduces the learning rate to a constant value that is smaller than the peak learning rate.
There are many annealing schedules, and we focus our discussion on the annealing schedules used by \ac{LLaMA-2} and \ac{T5}. 
\ac{LLaMA-2} uses a cosine annealing schedule, first proposed by \citet{Loshchilov2016SGDR:Restarts}, and \ac{T5} uses a learning rate of $\frac{1}{\sqrt{\max(n, k)}}$, where $n$ is the current training iteration and $k$ is the number of warm-up steps, which the authors set to $10^4$.
In their most recent work, \citep{Iyer2023Wide-minimaSchedule} discuss the \emph{wide minima density hypothesis}, which suggests that a high learning rate in the early stages of training increases the probability of the model to explore and arrive at areas with a high density of wide minima.
The authors propose a `knee-shaped' \emph{explore-exploit} learning rate schedule. Specifically, they show that training at a high learning rate (\emph{explore} phase) for an extended period of time before linearly decaying to zero (\emph{exploit} phase) yields higher performing models as it increases the probability of the model converging to a wide minima \citep{Iyer2023Wide-minimaSchedule}, which has been shown to lead to better generalization results \citep{Keskar2016OnMinima}.

All in all, we see that the literature on determining the warm-up scheme, peak learning rate, and annealing schedule varies widely and is largely based on empirical findings.
For the purposes of our research, we chose to adhere closely to established and empirically validated models to limit the extent of experimental variables we modify.

\paragraph{Batch sizes.}
As \ac{DL} models are trained with \ac{SGD}, a decision on the batch size has to be made.
As the batch size gets larger, gradients are more accurate and there is less noise in the gradients.
In the extreme case of batch gradient descent, where gradients are computed over the entire dataset, convergence to a local minimum is guaranteed but the performance of the model might be poor.
Therefore, \ac{SGD} is commonly used to introduce noise in the optimization process as it allows the escaping of local minima and potentially enables convergence to better minima.
The literature on determining suitable batch sizes has evolved over the past few years.

Until recently, it was widely accepted that large batch sizes result in a large generalization gap (or high test error).
\citet{Keskar2016OnMinima} first suggested that large batch sizes (512 and above) tend to converge to sharp minimizers \citep[cf., Figure 1,][]{Keskar2016OnMinima}, which results in a model with poor generalization and small batch sizes converge to flat minimizers \citep[cf., Figure 1,][]{Keskar2016OnMinima}, which results in a model with better generalization.
However, \citet{Hoffer2017TrainNetworks} show that it is not the batch size that affects model generalization, but rather the reduction in the number of \ac{SGD} updates that results from increasing the batch size while keeping the number of training epochs constant.

\citet{Goyal2017AccurateHour} show that large batch sizes can still result in a small generalization gap.
They successfully train ImageNet \citep{Deng2009ImageNet:Database}, a canonical \ac{CV} dataset for various \ac{CV} tasks, within one hour with a simple heuristic that scales the initial or peak learning rate by $k$, if the batch size increases by $k$.
This finding corroborates with an earlier technical report by \cite{Krizhevsky2014OneNetworks}, who suggests that the learning rate should be scaled by $\sqrt{k}$ as it scales the learning rate proportionately to the reduction in the standard deviation of the gradient estimator in a batch.
In practice, both $k$ and $\sqrt{k}$ are commonly used. 
We opted for $\sqrt{k}$ as a more conservative estimate as it is a smaller value than $k$.

\section{Retraining \ac{AM}}
\label{app:retraining-am}
We follow the default training parameters for \ac{AM} by training the models for 100 epochs, with each epoch having 1.28M training samples and utilizing the maximum available computational resources as prescribed by their code.
Specifically, their code is designed to operate on a single node and utilizes the maximum number of GPUs available on that node.
On our infrastructure, this translates to 2 Tesla V100-PCIE-32GB GPUs.
Additionally, we increase the batch size to its maximum limit without encountering an out of memory error on the GPUs, maintaining a consistent effective batch size of 512 for problem instances of sizes 20, 50, 100, and 200.
However, as the size of the problem instances grow, GPU memory constraints necessitate a reduction in batch size.
Consequently, for problem instance sizes of 400, 600, and 800, the batch sizes we employ are 128, 64, and 32, respectively.
Finally, the models for problem instances of sizes 20, 50, 100 and 200 successfully converged after 100 epochs, while the models for problem instances of sizes 400, 600 and 800 diverged within the first 10 epochs and thus we omit these results when comparing \ac{FM-MCVRP} against \ac{AM}.

\section{Test Statistics}
\label{app:test-stats}
We tested the distance distributions obtained in Section \ref{sec:outperform_heuristic} and Section \ref{sec:distance_distribution} with a one-sided paired samples t-test \citep{Ross2017PairedT-Test} as the test instances are the same.
In Section \ref{sec:outperform_heuristic}, the null hypothesis $H_0$ is testing if the solution values of \ac{FM-MCVRP} (\ac{NS}, 1,000 samples) is greater than or equal to the solution values of \ac{HGS} (single run) on 1,000 instances of a 400-customer problem in a paired samples t-test.
Given this null hypothesis, the alternative hypothesis $H_1$, is the direct opposite, where the solution values of \ac{FM-MCVRP} (\ac{NS}, 1,000 samples) is less than the solution values of \ac{HGS} (single run).
The one-sided paired samples t-test had a p-value of 0, allowing us to reject the null hypothesis.
This implies that \ac{FM-MCVRP} (\ac{NS}, 1,000 samples) has a higher performance in terms of solution value when compared with \ac{HGS} (single run).
Table \ref{table:test-stats-1} shows the details of the statistics.

\begin{table}[htbp]
\centering
\footnotesize
\begin{tabularx}{0.6\textwidth}{|X|c|c|}
\toprule
                & X & Y \\
\midrule
$H_0$              & \multicolumn{2}{c|}{$X \geq Y$} \\
$H_1$              & \multicolumn{2}{c|}{$X < Y$} \\
mean               & 34.71 & 35.08 \\
std                & 1.37 & 1.37 \\
t-stat             & \multicolumn{2}{c|}{-53.43} \\
p-value            & \multicolumn{2}{c|}{0.00} \\
degrees of freedom & \multicolumn{2}{c|}{999} \\
95\% CI            & \multicolumn{2}{c|}{-0.39 -- -0.36} \\
\bottomrule
\multicolumn{3}{l}{
            \tiny 
            X: \ac{FM-MCVRP} (\ac{NS}, 1,000 samples) ; 
            Y: \ac{HGS} (single run);
            }  \\
        \multicolumn{3}{l}{
            \tiny 
            Metrics are reported over 1,000 instances
            }  \\
\bottomrule
\end{tabularx}
\caption{Test statistics of comparing \ac{FM-MCVRP} (\ac{NS}, 1,000 samples) with \ac{HGS} (single run).}
\label{table:test-stats-1}
\end{table}

In Section \ref{sec:distance_distribution}, the null hypothesis $H_0$ is testing if the solution values of \ac{HGS} (best of 1,000 runs) is greater than or equal to the solution values of \ac{FM-MCVRP} (\ac{NS}, 1,000 samples) on 1,000 instances of a 400-customer problem in a paired samples t-test.
Given this null hypothesis, the alternative hypothesis $H_1$, is the direct opposite, where the solution values of \ac{HGS} (best of 1,000 runs) is less than the solution values of \ac{FM-MCVRP} (\ac{NS}, 1,000 samples).
The one-sided paired samples t-test had a p-value of 0, allowing us to reject the null hypothesis.
This implies that \ac{HGS} (best of 1,000 runs) has a higher performance in terms of solution value when compared with \ac{FM-MCVRP} (\ac{NS}, 1,000 samples).
Table \ref{table:test-stats-2} shows the details of the statistics.

\begin{table}[htbp]
\centering
\footnotesize
\begin{tabularx}{0.6\textwidth}{|X|c|c|}
\toprule
                & X & Y \\
\midrule
$H_0$              & \multicolumn{2}{c|}{$X \geq Y$} \\
$H_1$              & \multicolumn{2}{c|}{$X < Y$} \\
mean               & 34.43 & 34.71 \\
std                & 1.35 & 1.37 \\
t-stat             & \multicolumn{2}{c|}{-62.84} \\
p-value            & \multicolumn{2}{c|}{0.00} \\
degrees of freedom & \multicolumn{2}{c|}{999} \\
95\% CI            & \multicolumn{2}{c|}{-0.28 -- -0.27} \\
\bottomrule
\multicolumn{3}{l}{
            \tiny 
            X: \ac{HGS} (best of 1,000 runs) ; 
            Y: \ac{FM-MCVRP} (\ac{NS}, 1,000 samples);
            }  \\
        \multicolumn{3}{l}{
            \tiny 
            Metrics are reported over 1,000 instances
            }  \\
\bottomrule
\end{tabularx}
\caption{Test statistics of comparing \ac{HGS} (best of 1,000 runs) with \ac{FM-MCVRP} (\ac{NS}, 1,000 samples).}
\label{table:test-stats-2}
\end{table}

\section{Example Solutions}
\label{app:example-solutions}

Figure \ref{fig:small_best_gap} shows an example solution for a 400-node \ac{MCVRP} problem instance with the best gap when comparing \ac{FM-MCVRP} (\ac{NS}, 1,000 samples) with \ac{HGS} (best of 1,000 runs).
Observe how \ac{FM-MCVRP} finds solutions where the routes are tightly clustered within an angle compared to \ac{HGS}.
Figure \ref{fig:large_best_gap} shows an example solution for an 800-node \ac{MCVRP} problem instance with the best gap when comparing \ac{FM-MCVRP} (\ac{NS}, 1,000 samples) with \ac{HGS} (best of 1,000 runs).
Again, observe how \ac{FM-MCVRP} finds solutions where the routes are tightly clustered within an angle compared to \ac{HGS}.
Most notably, in this example, \ac{FM-MCVRP} found a solution with an additional route but has a better solution value.
These visualizations give insight to the solution distribution that \ac{FM-MCVRP} has learnt.

\begin{figure}[htbp]
    \centering
    \begin{subfigure}{0.5\textwidth}
        \centering
        \includegraphics[width=\textwidth]{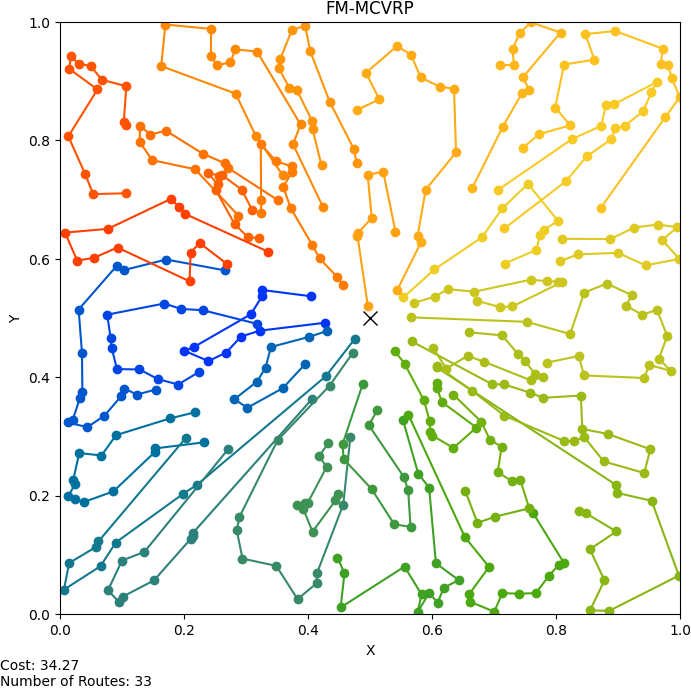}
        \captionsetup{font=scriptsize}
        \subcaption{\ac{FM-MCVRP} Solution}
        \label{fig:cvrp400_ours}
    \end{subfigure}%
    \hfill
    \begin{subfigure}{0.5\textwidth}
        \centering
        \includegraphics[width=\textwidth]{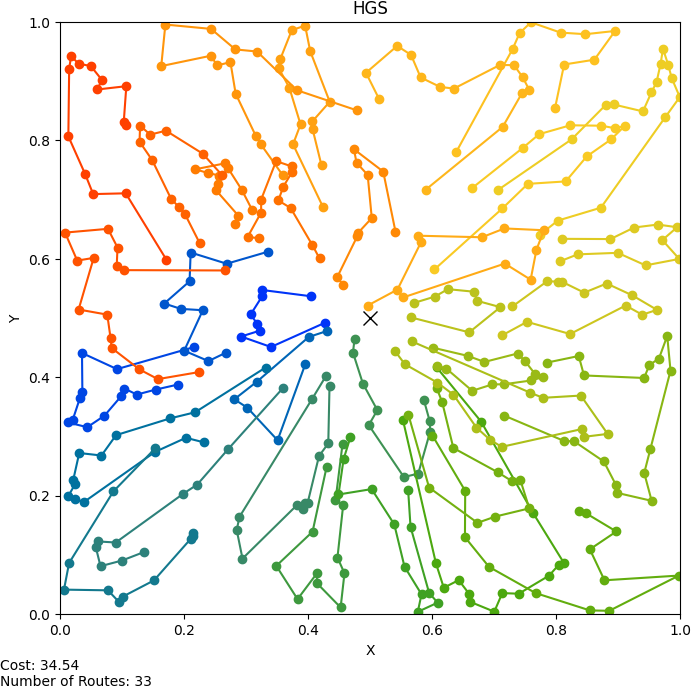}
        \captionsetup{font=scriptsize}
        \subcaption{\ac{HGS} Solution}
        \label{fig:cvrp400_hgs}
    \end{subfigure}%
    \caption{400-node problem instance with best gap when comparing \ac{FM-MCVRP} (\ac{NS}, 1,000 samples) with \ac{HGS} (best of 1,000 runs).}
    \label{fig:small_best_gap}
\end{figure}

\begin{figure}[htbp]
    \centering
    \begin{subfigure}{0.5\textwidth}
        \centering
        \includegraphics[width=\textwidth]{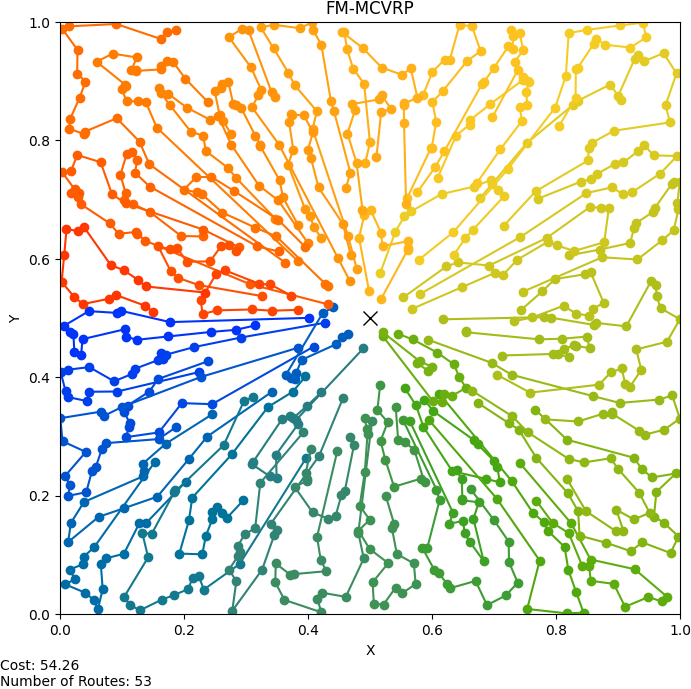}
        \captionsetup{font=scriptsize}
        \subcaption{\ac{FM-MCVRP} Solution}
        \label{fig:cvrp800_ours}
    \end{subfigure}%
    \hfill
    \begin{subfigure}{0.5\textwidth}
        \centering
        \includegraphics[width=\textwidth]{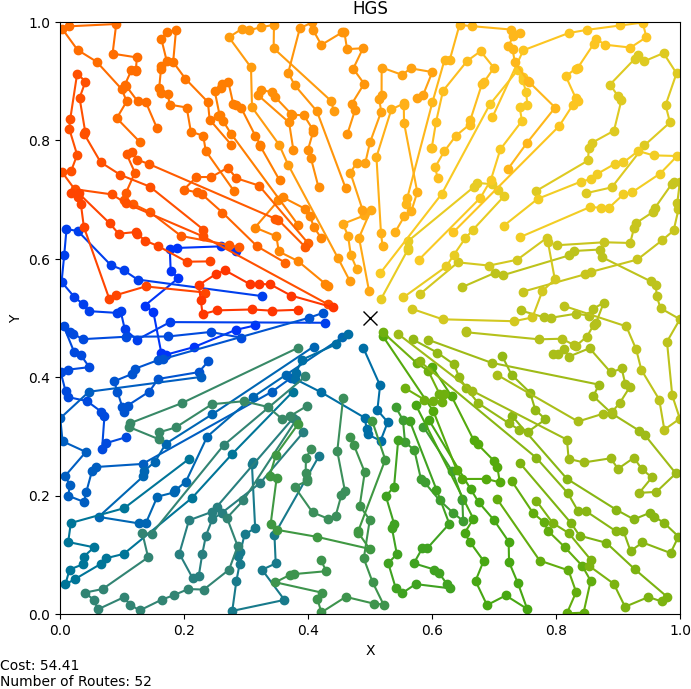}
        \captionsetup{font=scriptsize}
        \subcaption{\ac{HGS} Solution}
        \label{fig:cvrp800_hgs}
    \end{subfigure}%
    \caption{800-node problem instance with best gap when comparing \ac{FM-MCVRP} with \ac{HGS} on 1,000 samples. For this instance, observe how \ac{FM-MCVRP} found a shorter solution with more routes.}
    \label{fig:large_best_gap}
\end{figure}

\section{A Note on Route-Level Permutation Invariance}
The Transformer is a \ac{GNN} and \acp{GNN} have the properties of permutation invariance and permutation equivariance. 
A function $f$ is permutation invariant if $f(PX) = f(X)$, where $P$ is a permutation matrix and $X$ is the input matrix.
In each layer of our architecture, the features are matrices ($X$) and thus the order of the features can be permuted and the output will be the same.
Specifically, consider the routes produced by the solution section of our architecture (note that these are route IDs and not node IDs) $[0, 1, 2, 3, 4]$ and $[0, 1, 3, 2, 4]$. The embedding computed for the first node of route 4 is exactly the same and the order of the routes before that do not matter as the Attention mechanism transforms the input with a weighted sum and the sum operator is permutation invariant.
We hypothesized that this better represented the true space of solutions, and thus augmented the solutions during training by permuting our routes in this manner, giving the model multiple valid solutions for a given problem.
We experimented with multiple models trained in this manner and found that these models had extremely poor performance.
In particular, the routes were decoded in no particular order and was prone to miss out nodes in certain areas, which resulted in a giant loop being executed as the final route.
We found that ordering the solutions in a manner where routes that have similar angles with respect to the depot are grouped together alleviated the problem of having a giant loop.
Specifically, with reference to Figure \ref{fig:small_best_gap} or \ref{fig:large_best_gap}, an example of a solution that follows this order would have the blue routes decoded first and then sweeps counter clockwise to decode the green, yellow and eventually red routes.
While this specific ordering enabled us to achieve outperformance on \ac{HGS}, it remains an open question as to whether or not a permutation invariant operator can result in a higher performing model.

\end{APPENDICES}

\newpage


\ACKNOWLEDGMENT{The authors acknowledge the MIT SuperCloud and Lincoln Laboratory Supercomputing Center for providing HPC resources that have contributed to the research results reported within this paper.}






\bibliographystyle{informs2014}
\bibliography{references, jkschin} 

\end{document}